\documentclass[times,twocolumn,final]{elsarticle}
\usepackage{medima}
\usepackage{framed,multirow}

\usepackage{amssymb}
\usepackage{latexsym}

\usepackage{url}
\usepackage{xcolor}

\usepackage{hyperref}
\usepackage{subfigure}
\usepackage{bm}
\usepackage{multirow}
\usepackage{amsmath}
\usepackage{makecell}
\usepackage{booktabs}

\newcommand{\expect}[2]{\mathbb{E}_{#1}[#2]}

\definecolor{newcolor}{rgb}{.8,.349,.1}

\journal{Medical Image Analysis}

\begin{document}

\verso{Given-name Surname \textit{et~al.}}

\begin{frontmatter}

\title{BayeSeg: Bayesian Modeling for Medical Image Segmentation with Interpretable Generalizability\tnoteref{tnote1}}%
\tnotetext[tnote1]{This work was supported by the National Natural Science Foundation of China (61971142, 62111530195 and 62011540404) and the development fund for Shanghai talents (2020015).}

\author[1]{Shangqi \snm{Gao}\fnref{fn1}}
\author[1]{Hangqi \snm{Zhou}\fnref{fn1}}
\author[1]{Yibo \snm{Gao}\fnref{fn1}}
\author[1]{Xiahai \snm{Zhuang}\corref{cor1}}

\fntext[fn1]{These authors contributed equally to this work.}
\cortext[cor1]{Corresponding author: www.sdspeople.fudan.edu.cn/zhuangxiahai/}

%\ead[url]{www.sdspeople.fudan.edu.cn/zhuangxiahai/}

\address[1]{School of Data Science, Fudan University, Shanghai, 200433, China}

\received{1 March 2023}
\finalform{**}
\accepted{**}
\availableonline{**}
\communicated{**}

\begin{abstract}
%%%
%[\textbf{rewrite}] 
%Background
Due to the cross-domain distribution shift aroused from diverse medical imaging systems, many deep learning segmentation methods fail to perform well on unseen data, which limits their real-world applicability. 
Recent works have shown the benefits of extracting domain-invariant representations on domain generalization.
However, the interpretability of domain-invariant features remains a great challenge.
%%method部分
To address this problem, we propose an interpretable Bayesian framework (BayeSeg) through Bayesian modeling of image and label statistics to enhance model generalizability for medical image segmentation.
%which jointly models image and label statistics, utilizing the domain-independent shape of a medical image for segmentation. 
Specifically, we first decompose an image into a spatial-correlated variable and a spatial-variant variable, assigning hierarchical Bayesian priors to explicitly force them to model the domain-stable shape and domain-specific appearance information respectively. 
Then, we model the segmentation as a locally smooth variable only related to the shape. 
Finally, we develop a variational Bayesian framework to infer the posterior distributions of these explainable variables. The framework is implemented with neural networks, and thus is referred to as deep Bayesian segmentation.
%%results部分
Quantitative and qualitative experimental results on prostate segmentation and cardiac segmentation tasks have shown the effectiveness of our proposed method.
Moreover, we investigated the interpretability of BayeSeg by explaining the posteriors and analyzed certain factors that affect the generalization ability through further ablation studies. 
Our code will be released via
\href{https://zmiclab.github.io/projects.html}{https://zmiclab.github.io/projects.html}, once the manuscript is accepted for publication.
%and further ablation studies have explained the benefits of the explicit prior modeling and the joint modeling of image and label statistics on promoting generalization capability.
%%%%
\end{abstract}

\begin{keyword}
\MSC 41A05\sep 41A10\sep 65D05\sep 65D17
%% Keywords
\KWD Image segmentation\sep Interpretation and generalization\sep Statistical modeling\sep Variational Bayes
\end{keyword}

\end{frontmatter}

%\linenumbers

\section{Introduction}

%\textbf{[clinical background of medical image segmentation]} 
Medical image segmentation is aimed at classifying anatomical structures of different organs based on medical imaging techniques. Modern imaging techniques could provide diverse clinical signs, which include comprehensive pathological and structural information \citep{Hussain/2022}. However, new challenges for medical image segmentation are raised. Concretely, the intensity of medical images from different sequences, modalities, and sites can vary greatly due to potential factors in specific imaging systems \citep{dataset/MSCMR/2019,dataset/MMWHS/2016,dataset/ACDC/2018}, such as scanners, field-of-view, spatial resolution, signal-to-noise ratio, and digital processing software. These factors can result in distribution shifts among medical images \citep{rw/learning-meta-Dou/2019}, which makes cross-sequence, cross-modality, and cross-site segmentation particularly challenging.
%they raise a new challenge in medical image analysis, which is known as distribution shift. Concretely, appearances of two modalities could be greatly different; images from different sites vary necessarily; and multiple sequences exist for MRI. 
Although recent deep learning methods have achieved promising performance in computer vision and medical image analysis \citep{method/unet/2015}, they are vulnerable to the intrinsic distribution shift. That is, deep neural networks (DNNs) trained on one data set often cannot generalize well on the other unseen data set. 
Besides, collecting amounts of labeled medical images for DNNs to prevent over-fitting is expensive due to much labor of qualified experts.
%, but manually labeling medical images requires much labor of qualified experts. 
Therefore, pursuing generalizable models trained on small data sets becomes one of the key problems in medical image segmentation. 
To explore generalizable methods, a new field, known as domain generalization, has gained increasing attention in computer vision and medical image analysis \citep{rw/feature-alignment-adversarial01/2018,rw/learning-meta-Dou/2019}.

Current methods in domain generalization can be categorized into three groups, i.e., data-based, learning-based, and representation-based \citep{rw/survey_DG/2022}. The data-based methods try to enrich the diversity of training data by data augmentation or data generalization \citep{rw/data-augmentation-adversarial01/2018,rw/data-augmentation-adversarial03/2022, rw/data-generation-Randconv/2020}. 
These methods are effective for small distribution shifts but can be vulnerable to significant domain gaps. 
The learning-based methods aim to combine knowledge from training domains by specific strategies, including ensemble learning \citep{rw/learning-ensemble01/2019, rw/learning-ensemble02/2021} and meta-learning \citep{rw/learning-meta-Dou/2019, rw/learning-meta-shape/2020}. 
These methods have achieved promising performance on specific target domain, but limiting the number of source domains could weaken their generalization ability.
%These methods require online learning on the test domain, which is inefficient in practical applications due to long inference time.
The representation-based methods are developed to extract domain-invariant features through feature alignment \citep{rw/feature-alignment-adversarial01/2018,rw/feature-alignment-entropy/2020,rw/feature-alignment-kernel/2013}  or feature disentanglement \citep{rw/feature-disentanglement-diva/2020, rw/feature-disentanglement-causality/2021}. 
Current representation-based approaches are devoted to representing domain-invariant information through aligned hidden feature maps. Although these methods deliver promising generalization ability in computer vision, little attention has been paid to model interpretability, intuitively, the ability to provide explanations in understandable terms to human experts \citep{back/interpretability_survey/2021}. It could raise concerns on ethical and legal requirements in clinical diagnosis and treatment \citep{Zhang/2023}. 
Therefore, exploring interpretable and generalizable representations is more attractive for medical image segmentation.

\begin{figure*}[t]
    \centering
    \includegraphics[width=0.95\linewidth]{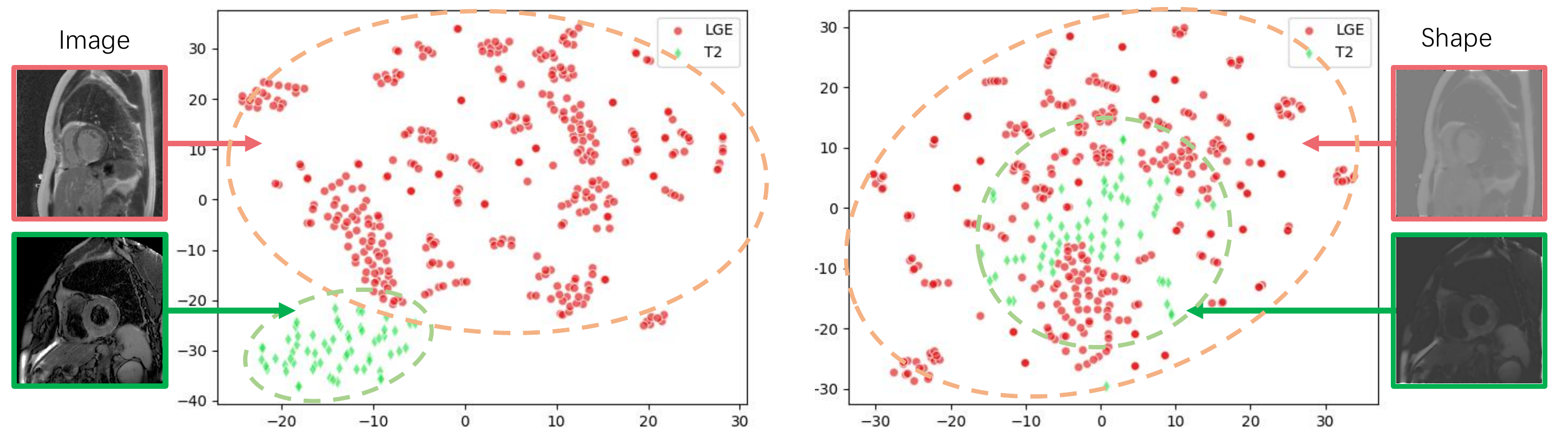}\\[-3ex]
    \caption{\textbf{Left: }t-SNE visualization of the cross-sequence domain shift in medical image segmentation. Conventional deep learning methods which depends on domain-variant features (e.g. appearance) to segment could fail to generalize on the unseen domains. \textbf{Right:} t-SNE visualization of the decomposed shape of two different domains, whose distributions stay close through the explicit statistical modeling. Hence, segmentation on shape information would lead to the promoted generalizability.}
    \label{fig:tsne-sample}
\end{figure*}
%\textbf{[our methods with statistical modeling]} 

 Statistical modeling is able to improve model interpretability and generalization capability \citep{Gao/BayeSR/2023}. 
 Motivated by this, instead of implicit representation, we explore explicit modeling for shape information, which represents structures of organs.
%Motivated by this, we extract shape, which represents structures of organs, by image decomposition, and jointly describe statistics of images and labels for medical image segmentation. 
 As Fig. \ref{fig:tsne-sample} shows, there is a significant distribution shift between two different MR sequences, but such distribution shift can be well reduced by extracting shape information. 
That is, the extracted shape is more likely to be a domain-invariant representation and segmenting from the shape has the potential of improving generalizability. 
Therefore, we are going to jointly characterize image and label statistics to extract interpretable shape representation for generalizable segmentation.
%Therefore, we are going to develop a deep learning framework for medical image segmentation by combing statistical modeling with image decomposition.
%Therefore, instead of implicit representation, we explore an explicit statistical representation for shape information.
% Therefore, we are going to joint model of image and label statistics to encourage model to extract understandable shape representations.

%\textbf{[our Bayesian framework]} 
In this work, we propose an interpretable and generalizable Bayesian segmentation (BayeSeg) framework by joint modeling image and label statistics. 
%To the best of our knowledge, this is the first attempt of combining image decomposition, image segmentation, and deep learning. 
Concretely, we first represent an image as two variables. 
One is spatial-correlated for representing shape information, and the other is spatial-variant for representing appearance information. Since both variables are unconstrained, we further assign hierarchical Bayesian priors to explicitly describe their statistics.
Afterward, we utilize variational inference to approximate their posteriors due to intractable computation.
Finally, we build a deep neural networks to implement the variational method and infer the posterior distributions of the shape, appearance, and segmentation from given images.

%\textbf{[our contributions]} 
This work is extended from our conference paper on MICCAI 2022 \citep{method/bayeseg_miccai/2022}, in which we propose a new Bayesian segmentation framework (BayeSeg) by joint modeling of image and label statistics, and solve the model through a variational Bayesian approach implemented by neural networks. 
Our new contributions are listed as follows:
\begin{itemize}
    \item We further improve the formulation of BayeSeg by statistically describing the relationship between segmentation and its ground-truth.
    \item We conduct extensive experiments on cardiac and prostate segmentation, which covers cross-modality, cross-sequence and cross-site settings. Besides, we clarify the clinical background of distribution shift and review the development of domain generalization.
    \item We elaborate on the model interpretability by visualizing and explaining posteriors extracted from BayeSeg. Furthermore, we investigate the model generalizability by selecting statistical priors, pruning probabilistic model, and analyzing data distribution.
\end{itemize}

The rest of our paper is organized as follows. 
In Section 2, we introduce and summarize the related literature on statistical modeling and domain generalization. 
In Section 3, we present the methodology, including formulation, network architectures and training strategies. 
Section 4 provides the experimental settings and validates the generalizability of BayeSeg.
% Section 4 provides the implementation details of BayeSeg and the experimental generalization results on three segmentation tasks over different organs (cardic, abdomen, prostate). 
In Section 5, we study the interpretability and generalizability of the proposed method. 
Section 6 discusses the limitations of BayeSeg and our future works. Finally, the conclusion of this work is drawn in Section 7.

\section{Related works}

\subsection{Image statistical modeling}
% The framework of spatial regression can be considered as 
% \begin{equation}
%     y = X\beta + z + \epsilon,
% \end{equation}
% where, the first term model a first-order mean structure that includes covariates with regression coefficients $\beta$; the second term denotes a latent spatial random error following $\mathcal N(z|0, \Sigma)$; and the third term denotes an independent error following $\mathcal N(\epsilon| 0, \sigma_{\epsilon}) I$. 

% The spatial regression framework becomes a simultaneous autoregressive model when $\Sigma$ takes a form of:
% \begin{equation}
%     \Sigma = \sigma_z^2 ((I - B)(I - B^T))^{-1},
% \end{equation}
% and a conditional autoregressive model when $\Sigma$ takes a form of:
% \begin{equation}
%     \Sigma = \sigma_z^2 (I - C)^{-1} M.
% \end{equation}
% Here, $B$ and $C$ model spatial dependence between any two elements of $z$ for SAR and CAR, respectively. Particularly, the \mathop{\mathrm{diag}}onal elements of $B$ and $C$ are zeros, and $M$ is a \mathop{\mathrm{diag}}onal matrix. The spatial dependence matrices are often expressed as $B = \rho W$ and $C = \rho W$, where $W$ is a weight matrix indicating neighbor relationships and $\rho$ controls the strength of dependence. 

Statistically describing the similarity between intensity values of pixels is important in image processing. The resulting image priors have been widely used in image restoration \citep{Gibbs/restoration/1994}, image classification \citep{MRF/classification/1996}, and image segmentation \citep{MRF/segmentation/2000}. 
To avoid computational intractability, early works focused on \textbf{Gaussian prior modeling}. \cite{Hunt/1977} used multivariate Gaussian probability density to model noise and intensity of real-world images by setting covariance to be a Toeplitz matrix. \cite{Archer/1995} also adopted Gaussian prior for restored images by setting inverse covariance to be a non-negative definite matrix. Although these models are tractable, they cannot sufficiently describe the relationship among pixels.
To address this problem, much attention has been paid to \textbf{image field modeling}. For describing a textured image, \cite{Cross/1983} considered a texture to be a stochastic image field, and explored the use of Markov random field (MRF). \cite{Geman/1984} regarded an image as a pair of variables following MRF, where one was used to model the intensity of pixels, while the other was used to detect edges. \cite{Gibbs/restoration/1994} modeled a class of images with piecewise homogeneous regions by defining Gibbs distribution, which is equivalent to constructing MRF. \cite{Li/1996} used a Markov random field to model contextual interactions of neighboring sites of multi-level images or segmented maps. To avoid the smoothing of edges, \cite{Pan/2006} combined a Huber function with an MRF to model image priors. These models are effective in characterizing image statistics, but are challenging in computation due to the dependence among all pixels. To pursue tractable modeling, some works were devoted to \textbf{spatial correlation modeling}. \cite{Mesarovic/1998} and \cite{Molina/1999} used a simultaneous autoregressive (SAR) prior coupled with a circulant Laplacian high-pass operator to control the smoothness of images. After that, \cite{Molina/2003} adopted a conditional autoregressive (CAR) prior to describe the smoothness of objects, and introduced the concept of line to consider edges between two neighbor pixels. The aforementioned methods all focused on statistical modeling in the image domain. To explore statistics in the transform domain, some works were aimed at \textbf{sparsity modeling}. Instead of directly modeling images, \cite{Portilla/2003} used a Gaussian scale mixture model to describe the statistical relationship between wavelet coefficients of an image. To preserve image edges, \cite{Chantas/2006} modeled the first-order differences of an image by Gaussian random variables with zero mean and spatially varying variance, and further imposed the variance on a Gamma hyper-prior. \cite{Portilla/2015} described the sparsity under a series of linear transforms based on maximum entropy analysis. The previous methods all used fixed operators to capture the difference between neighboring pixels, which could limit the flexibility of statistical modeling. To overcome this limitation, a few works contributed to \textbf{flexible prior modeling}. Instead of using first-order MRF, \cite{Zhang/2012} used a high-order MRF to capture the statistics of natural images by learning the high-order filters from data. \cite{Gao/2018} also defined a high-order MRF based on Gaussian scale mixtures with learnable linear filters and weights. 
Inspired by the success of these works, we are going to model statistics of medical images in this work. Since the intensity values of medical images can be the composition of spatially independent and spatially correlated information, we will use spatially variant Gaussian priors to model the spatial independence, while adopting SARs to model the spatial correlation.

\subsection{Domain generalization for medical image segmentation}
The aim of domain generalization (DG) is to make a model trained on one or several source domains generalize well on unseen target domains.
Different from domain adaptation (DA), DG has no access to target domain data during the training process. This makes DG more challenging, but closer to real-world situations like clinical applications. 
In general, 
%from the perspective of the number of source domains, DG approaches can be assigned to multi-source domain generalization (MDG) and single-source domain generalization (SDG). 
from the perspective of techniques, DG methods can be roughly categorized into three groups, i.e., data-based approaches, learning-based approaches and representation-based approaches \citep{rw/survey_DG/2022}.
\textbf{Data-based approaches} \citep{rw/data-augmentation-cutout/2017, rw/data-feature-DSU/2022} enrich the diversity of data distribution in training by data augmentation or data generation to enhance model generalization ability. For medical image segmentation, 
%\cite{rw/data-generation-01/2020} synthesized examples from pre-built generative models to generalize on unseen domains. 
\cite{rw/medical-Zhang/2020} applied a series of stacked transformations to simulate domain shift for a specific medical imaging modality.
\cite{rw/data-generation-AdvBias/2020} learned the bias-field and deformation field to generate plausible and realistic signal corruptions through adversarial learning, and
%\cite{rw/data-generation-Challengesample/2021} masked the decoupled latent space in both channel-wise and spatial-wise manners to generate challenging samples.
\cite{rw/data-augmentation-adversarial03/2022} adapted adversarial data augmentation, and proposed a mutual information regularizer to promote the cross-domain semantic consistency.
\cite{rw/feature-disentanglement-causality/2022} proposed a causalty-inspired data augmentation method to simulate different possible MRI imaging processes for domain-invariant segmentation networks.
%newly developed techniques include adversarial data augmentation \cite{rw/data-augmentation-adversarial01/2018,rw/data-augmentation-M-ADA/2020, rw/data-augmentation-adversarial03/2022}, image texture randomization \cite{rw/data-generation-Randconv/2020}), challenging sample generation \cite{rw/data-generation-Challengesample/2021, rw/data-generation-AdvBias/2020}, style transfer \cite{rw/data-generation-mixstyle/2021}, dynamic networks \cite{rw/data-generation-PDEN/2021} and so on. 
\textbf{Learning-based approaches} focus on improving the learning strategy. \cite{rw/learning-meta-Dou/2019} adapted meta-learning for generalization, and introduced two complementary losses to encourage shape compactness and shape smoothness.
\cite{rw/learning-meta-shape/2020} further considered the boundary of prediction masks, proposing a shape-aware meta-learning scheme SAML for prostate MRI segmentation. 
\cite{rw/learning-meta-disentanglement/2021} explicitly modeled and disentangled the representations to better approximate the domain shifts for meta-learning.
Besides, \cite{rw/medical-FedDG/2021} proposed an episodic learning strategy in continuous frequency space, which balances and fuses the information from multi-source domains to build a general model.
%Meta-learning methods \cite{rw/learning-Episodic/2019, rw/learning-meta-Dou/2019,rw/learning-meta-shape/2020,rw/learning-meta-disentanglement/2021} simulate domain shift to learn general representations. Ensemble learning \cite{rw/learning-ensemble01/2019,rw/learning-ensemble02/2021} methods balance and fuse the information from multi-source domains to build a general model. Besides, \cite{rw/learning-self/2019} promotes model generalizability by self-supervised learning. \cite{rw/learning-gradient/2019} proposed a self-challenging generalization method by removing the dominant features based on gradients. 
\textbf{Representation-based approaches} are devoted to extracting domain-invariant or domain-irrelevant features for the model to make decisions. Specifically, feature alignment methods aim to narrow the cross-domain feature distribution gap to make features domain-invariant. %by kernel mapping\cite{rw/feature-alignment-kernel/2013}, adversarial learning \cite{rw/feature-alignment-adversarial02/2018, rw/feature-alignment-adversarial01/2018, rw/feature-adversrial-Zhao/2021}, or explicitly minimizing the distribution distance as regularization terms \cite{rw/feature-alignment-adversarial01/2018, rw/feature-alignment-entropy/2020}. 
\cite{rw/feature-alignment-adversarial02/2018, rw/feature-alignment-adversarial01/2018, rw/feature-adversrial-Zhao/2021} adapted adversarial learning to align the distributions among different domains. To explicitly minimizing the distribution distance.  \cite{rw/feature-alignment-adversarial01/2018} also imposed the Maximum Mean Discrepancy (MMD) measure, and \cite{rw/feature-alignment-entropy/2020} proposed an entropy regularization term for conditional distributions. 
Feature disentanglement methods aim to decompose a feature representation into domain-specific parts and domain-shared/irrelevant parts. %Common implementations include VAE framework\cite{rw/feature-disentanglement-diva/2020} and causality-inspired models \cite{rw/feature-disentanglement-causality/2021, rw/feature-disentanglement-causality/2022}, which focus on the causal relationship between latent features and outputs to find the most important features that cause the results.
For instance, \cite{rw/feature-disentanglement-diva/2020} disentangled the image features into domain-related, class-related, and residual-related variables for classification, and learned those independent latent subspaces through a domain-invariant variational auto-encoder (DIVA). 
Recently, causality-inspired models \citep{rw/feature-disentanglement-causality/2021, rw/feature-disentanglement-causality/2022} focus on the causal relationship between latent features and outputs to find the most important features that cause the results.
Although these representation-based methods have achieved remarkable progress in many computer vision tasks, there are relatively few of them targeting medical image segmentation \citep{rw/feature-alignment-kernel/2013, rw/feature-adversrial-Zhao/2021, rw/feature-disentanglement-causality/2022}. 
Moreover, the interpretability of the domain-invariant features themselves remains a great challenge to promote. 

To further enhance the interpretability of features, we combine statistical modeling with image decomposition and achieve generalizable segmentation through a Bayesian framework. Unlike DSU \citep{rw/data-feature-DSU/2022} that models domain shifts with the uncertain feature statistics for data augmentation, we aim to extract the shape information that represents the domain-invariant organ structures. Different from \cite{rw/feature-alignment-adversarial02/2018} and \cite{rw/feature-alignment-entropy/2020}, which rely on multiple source domains for feature alignment, BayeSeg constrains the posteriors of features through assigning hierarchical Bayesian priors. Furthermore, our approach diverges from DIVA \citep{rw/feature-disentanglement-diva/2020}, which disentangles images into an uninterpretable low-dimensional representation for classification, as we focus on explicit statistical modeling to yield an interpretable representation in medical image segmentation.

\section{Methodology}
\begin{table}[!t]
\caption{Summary of mathematical notions and corresponding notations. Here, VDs denote variational distributions.}\label{tab1:notation}
\centering
\resizebox{\linewidth}{!}{
\begin{tabular}{|l|l|}
\hline
Notation & Notion\\
%\hline
%Scalar & lowercase letter, e.g., $a$\\
%Vector & boldface lowercase letter, e.g., $\bm a$\\
%Notion & boldface capital letter, e.g., $\bm D$\\
\hline
$\bm y\in \mathbb{R}^{d_{y}}$ & Image \\
$\bm u \in \mathbb{R}^{d_{y} \times K}$ & Ground-truth Label \\
%Domain-independent shape
\hline
$\bm x \in \mathbb{R}^{d_{y}}$ & Shape\\
%Domain-related appearance 
$\bm a \in \mathbb{R}^{d_{y}}$ & Appearance\\
%Softmax of segmentation logits 
$\bm z \in \mathbb{R}^{d_{y} \times K}$ & Segmentation\\
\hline
$\bm \upsilon \in \mathbb{R}^{d_{y}}$ & Shape boundary\\
$\bm m \in \mathbb{R}^{d_{y}}$ & Appearance mean\\
$\bm \rho \in \mathbb{R}^{d_{y}}$ & Appearance inverse variance\\
$\bm \omega \in \mathbb{R}^{d_{y} \times K}$& Segmentation boundary\\
$\bm \pi \in \mathbb{R}^{K}$ & Label portion\\
\hline
$\mathcal{N}(\cdot, \cdot)/ \mathcal{G}(\cdot, \cdot)/ \mathcal{B}(\cdot, \cdot)$& Gaussian/Gamma/Beta distribution\\
$\bm \mu_{\cdot}^{(0)}$, $\bm \sigma_{\cdot}^{(0)}$, $\bm \phi_{\cdot}^{(0)}$, $\bm \gamma_{\cdot}^{(0)}$, $\bm \alpha_{\cdot}^{(0)}$, $\bm \beta_{\cdot}^{(0)}$ & Hyperparameters of distributions\\
% $\mathcal{G}(\cdot,\cdot)$ & Gamma distribution\\
% $\bm \phi_{\cdot}^{(0)}$, $\bm \gamma_{\cdot}^{(0)}$& Hyperparameters of Gamma distribution\\
% $\mathcal{B}(\cdot, \cdot)$ & Beta distribution\\
% $\bm \alpha_{\cdot}^{(0)}$, $\bm \beta_{\cdot}^{(0)}$& Hyperparameters of Beta distribution\\
$\bm \hat{\mu}_{\cdot} / \bm \hat{\sigma}_{\cdot}$ & Mean/Deviation of VDs\\
\hline
\end{tabular}
}
\end{table}

\begin{figure*}[!t]
    \centering
    \subfigure[Probabilistic Graphical Model]{\includegraphics[width=0.27\linewidth]{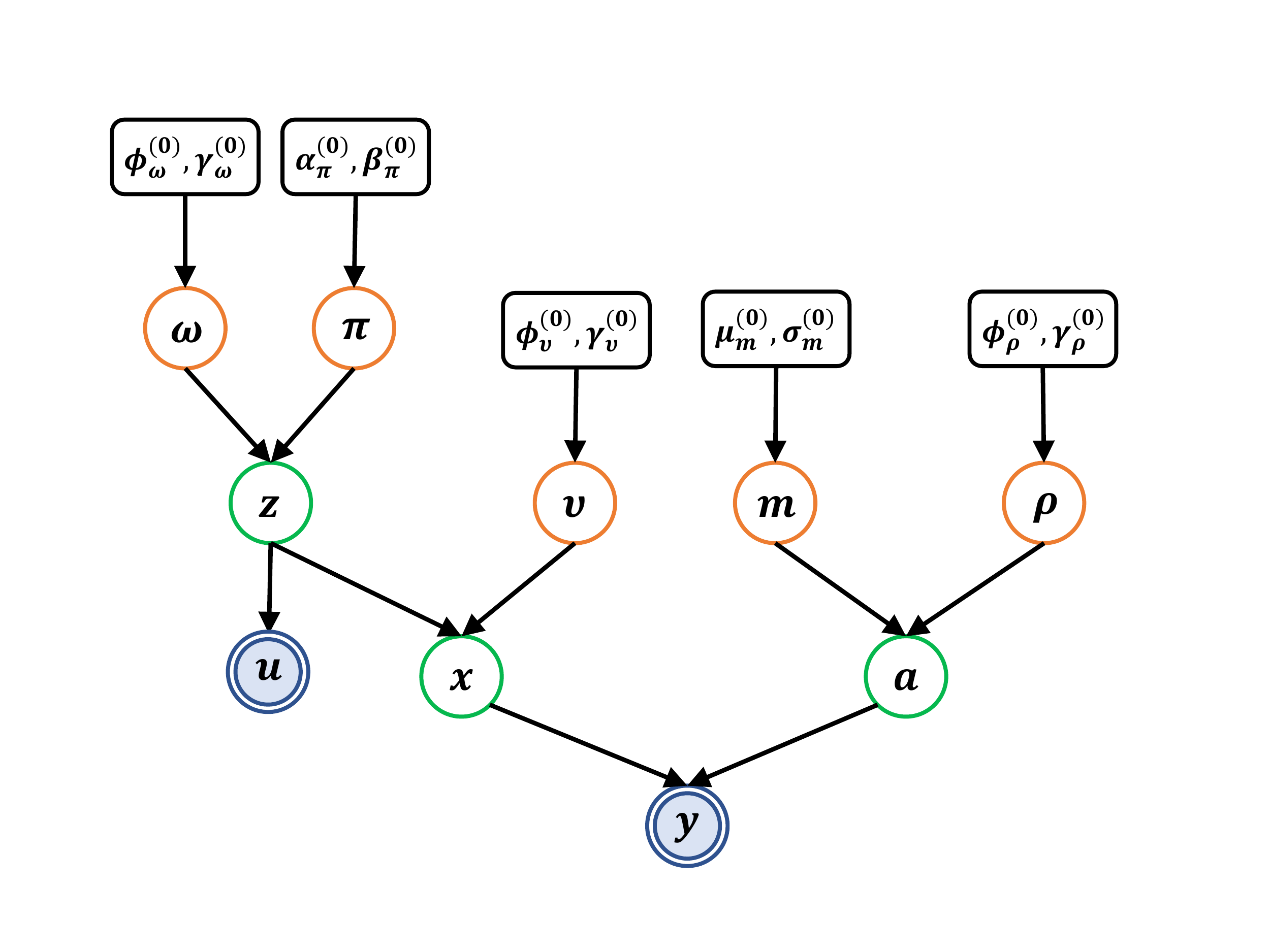}}
    \subfigure[Deep learning framework]{\includegraphics[width=0.72\linewidth]{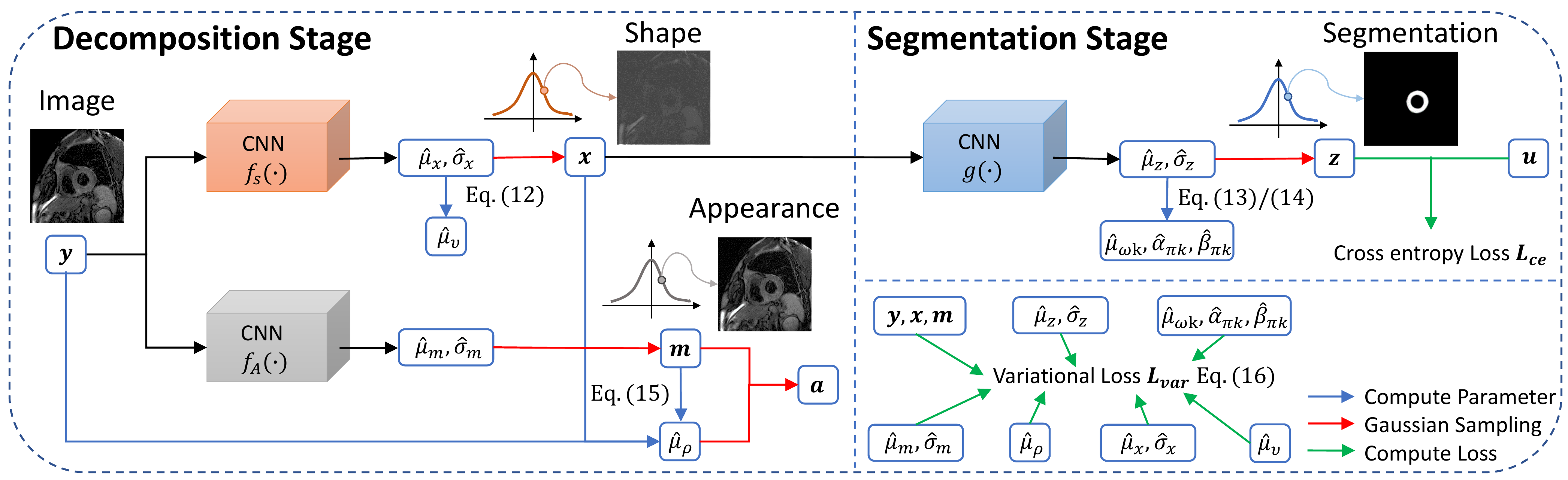}}
    \caption{The framework of Bayesian segmentation (BayeSeg). (a) shows the probabilistic graphical model (PGM) of BayeSeg. Here, blue circles denote the observed image and its corresponding ground truth label, green and orange circles denote unknown variables, and black rectangles denote hyper-parameters. (b) presents the deep learning framework of BayeSeg. Given an image, we first use two convolutional neural networks (CNNs) to infer the posterior distributions of the shape and appearance respectively, and obtain their random samples. Then, we use another CNN to infer the posterior distribution of segmentation, and the resulting random sample is the prediction in the training process. Please refer to Section \ref{sec:train} for the details of network architecture and training strategy.}
    \label{fig:framework}
\end{figure*}

%Many learning-based methods trained on one sequence MR images cannot generalize well to the other sequence or site data \cite{back/Li/2019}.
This work develops an interpretable Bayesian segmentation (BayeSeg) framework to enhance the generalizability of deep learning models. The framework mainly consists of two parts, i.e., (1) statistical modeling of images and labels, and (2) deep inference of posteriors. BayeSeg could be effective in enhancing model generalizability due to the joint modeling of image and label statistics. Although conventional variational Bayesian approaches \citep{back/Blei/2017} can be used to solve the statistical model induced from (1), they are computationally expensive in inferring high-dimensional variables due to thousands of online iterations. Inspired by the efficiency of DNN in learning complex mappings, we proposed to infer the posteriors by deep learning. For convenience, the notions and notations used in this paper are summarized in Table \ref{tab1:notation}.
%At the first stage, we build a probabilistic graphical model (PGM) for the modeling of image and label statistics. Specifically, we decompose an image into two variables representing its shape and appearance information, and only the shape is related to an expected segmentation. 
%At the second stage, we first build two CNNs to infer the posterior distributions of the shape and appearance, respectively. Then, we build a segmentation network to estimate the posterior distribution of the segmentation.  

Fig. \ref{fig:framework} shows the framework of statistical modeling and deep inference of the proposed BayeSeg. For the statistical modeling, we build a probabilistic graphical model (PGM) as shown in Fig. \ref{fig:framework} (a). Concretely, we first decompose an image $\bm y$ into two variables. One is the spatial-correlated shape $\bm x$ which models the information of anatomic structures, and the other is the spatial-variant appearance $\bm a$ which models the textural information of the image. 
Since the structure of organs is domain-irrelevant, the model predicting a label from the shape could have better generalizability. 
To hierarchically characterize the statistics of images and labels, the shape $\bm x$ is conditioned on its boundary $\bm \upsilon$ for detecting the edges of structures and the segmentation $\bm z$, while the appearance $\bm a$ is set to be a Gaussian variable determined by the spatial-variant mean $\bm m$ and the spatial-variant inverse variance $\bm \rho$. Similarly, the segmentation $\bm z$ is forced to be locally smooth and depends on its boundary $ \bm \omega $ and the label portion $\bm \pi$ of all classes. 
Besides, the ground-truth label $\bm u$ is modeled as a variable depending on the segmentation, which follows the Bernoulli distribution. Finally, Gamma priors are assigned to $\bm \rho$, $\bm \upsilon$, and $\bm \omega$, a Beta prior is assigned to $\bm \pi$, and a Gaussian prior is assigned to $\bm m$. 
For the deep inference, we build a deep learning framework as shown in Fig. \ref{fig:framework} (b). First, we use two CNNs to separately infer the posterior distributions of the shape and appearance. Second, we use a CNN to estimate the posterior distribution of the segmentation from the shape. Finally, to train these CNNs, the parameters of posteriors will be jointly used to compute a variational loss for maximum a \textit{posteriori}, which will be combined with a cross entropy loss for maximum label likelihood.

\begin{table}
\caption{Forms of conjugate priors and variational posteriors.} \label{tab:priors}
\centering
\resizebox{1\linewidth}{!}{
    \begin{tabular}{lll}
    \toprule
     Variable   & Conjugate prior   & Variational posterior    \\
     \toprule
     $\bm m$ & $\mathcal{N}(\bm m|\mu_{m}^{(0)}\bm 1, {(\sigma_{m}^{(0)}})^{-1}\bm I)$ & $\mathcal{N}(\bm m| \hat{\bm\mu}_{m}, \mathop{\mathrm{diag}}(\hat{\bm\sigma}_{m}^{2}))$ \\
     \midrule
     $\bm \rho$ & $\prod^{d_{y}}_{i=1}\mathcal{G}(\rho_{i}|\phi_{\rho i}^{(0)}, \gamma_{\rho i}^{(0)})$&$\prod^{d_{y}}_{i=1}\mathcal{G}(\rho_{i}|\hat{\phi}_{\rho i}, \hat{\gamma}_{\rho i})$\\
      \midrule
      $\bm \upsilon$ &
      $\prod^{d_{y}}_{i=1}\mathcal{G}(\upsilon_{i}|\phi_{\upsilon i}^{(0)}, \gamma_{\upsilon i}^{(0)})$&
      $\prod^{d_{y}}_{i=1}\mathcal{G}(\upsilon_{i}|\hat{\phi}_{\upsilon i}, \hat{\gamma}_{\upsilon i})$\\
      \midrule
      $\bm \omega$ &$\prod_{i=1}^{d_{y}}\prod_{k=1}^{K}\mathcal{G}(\omega_{ki}|\phi_{\omega ki}^{(0)}, \gamma_{\omega ki}^{(0)})$  &$\prod_{i=1}^{d_{y}}\prod_{k=1}^{K}\mathcal{G}(\omega_{ki}|\hat{\phi}_{\omega ki}, \hat{\gamma}_{\omega ki})$\\
      \midrule
      $\bm \pi$ &
      $\prod_{k=1}^{K}\mathcal{B}(\pi_{k}|\alpha_{\pi k}, \beta_{\pi k})$ &
      $\prod_{k=1}^{K}\mathcal{B}(\pi_{k}|\hat{\alpha}_{\pi k}, \hat{\beta}_{\pi k})$ \\
     \bottomrule
    \end{tabular}
}
\end{table}

\subsection{Statistical modeling of image and label}
Statistical modeling of images and labels aims at exploring their internal statistics by probabilistic methods, which is useful in medical image segmentation. Many previous works have shown that image statistics is important in image restoration \citep{Gibbs/restoration/1994}, image classification \citep{MRF/classification/1996}, and image segmentation \citep{MRF/segmentation/2000}, but modeling textural images is still challenging. Medical image segmentation not only requires to describe image statistics, but also needs to characterize label statistics. Therefore, we propose to jointly model image and label statistics for medical image segmentation.

For $K$-class segmentation, given an image sampled from the variable $\bm y \in \mathbb{R}^{d_{y}}$, where $d_{y}$ denotes the dimension of $\bm y$, we decompose $\bm y$ into the sum of a shape $\bm x$ and an appearance $\bm a$, i.e., $\bm y = \bm x + \bm a$.
Then, the appearance $\bm a$ is modeled to be spatial-variant, following a Gaussian distribution with a mean $\bm m \in \mathbb{R}^{d_{y}}$ and a covariance $\mathop{\mathrm{diag}}(\bm \rho)^{-1} \in \mathbb{R}^{d_{y} \times d_{y}}$, i.e., $ p(\bm a|\bm m,\bm \rho) = \mathcal{N}(\bm a|\bm m, \mathop{\mathrm{diag}}(\bm \rho)^{-1}) $. 
Therefore, the image likelihood can be expressed as
\begin{equation}\label{eq:likelihood}
    p(\bm y|\bm x,\bm m, \bm \rho) = \mathcal{N}(\bm x + \bm m, \mathop{\mathrm{diag}}(\bm \rho)^{-1}).
\end{equation}

%The appearance $\bm a$ of an image is determined by a Gaussian distribution, that is, $ p(\bm a|\bm m,\bm \rho) = \mathcal{N}(\bm a|\bm m, \mathop{\mathrm{diag}}(\bm \rho)^{-1}) $. Specifically, we assign a Gaussian prior to $\bm m$, i.e., $p(\bm m|\mu_{m}^{(0)}, \sigma_{m}^{(0)}) = \mathcal{N}(\bm m|\mu_{m}^{(0)}\bm 1, {(\sigma_{m}^{(0)}})^{-1}\bm I)$, and a Gamma prior to $\bm \rho$, namely, $p(\bm \rho|\bm \phi_{\rho}^{(0)},\bm \gamma_{\rho}^{(0)}) = \prod^{d_{y}}_{i=1}\mathcal{G}(\rho_{i}|\phi_{\rho i}^{(0)}, \gamma_{\rho i}^{(0)})$. Here, $\bm I$ denotes an identity matrix, $\mu_{m}^{(0)}, \sigma_{m}^{(0)}, \bm \phi_{\rho}^{(0)}$, and $\bm \gamma_{\rho}^{(0)}$ are predefined hyper-parameters, and $\mathcal{G}(\cdot,\cdot)$ represents the Gamma distribution.

The shape $\bm x$ is modeled to be spatial-correlated through a simultaneous autoregressive model (SAR) \citep{method/SAR/2008}. It depends on an expected segmentation $\bm z \in \mathbb{R}^{d_{y} \times K} $ and a boundary $\bm \upsilon \in \mathbb{R}^{d_{y}} $ indicating edges of the shape, namely,
\begin{equation}\label{eq:shape}
    p(\bm x|\bm z, \bm \upsilon)
    = \textstyle\prod_{k=1}^{K} \mathcal{N}(\bm x|\bm 0, [\bm D_{x}^{T}\mathop{\mathrm{diag}}(\bm z_{k}\bm\upsilon)\bm D_{x}]^{-1}),
\end{equation}
where, $\bm z_{k}$ denotes the segmentation of the $k$-th class; $\bm \upsilon$ indicates the edges of the shape; $\bm D_x = \bm I - \bm B_{x}$ is a non-singular matrix; and each row of $\bm B_{x} \in \mathbb{R}^{d_{y} \times d_{y}} $ describes the neighboring system of one central pixel. For example, the values of $\bm B_{x}$ for the nearest four pixels equal to -0.25 while others are zeros, and thus $\bm D_x $ is a Laplacian operator.

%and it is assigned a Gamma prior with hyper-parameters $\bm \phi_{\upsilon}^{(0)}$ and $\bm \gamma_{\upsilon}^{(0)}$, i.e., $p(\bm\upsilon|\bm \phi_{\upsilon}^{(0)},\bm \gamma_{\upsilon}^{(0)}) = \prod_{i=1}^{d_{y}}\mathcal{G}(\upsilon_{i}|\phi_{\upsilon i}^{(0)}, \gamma_{\upsilon i}^{(0)})$.

The segmentation $\bm z$ is modeled by another SAR depending on a segmentation boundary $\bm \omega \in \mathbb{R}^{d_{y} \times K}$ and a class-wise probability $\bm \pi \in \mathbb{R}^K$, namely,
\begin{equation}\label{eq:label}
    p(\bm z|\bm \pi, \bm \omega) = \textstyle\prod_{k=1}^{K} \mathcal{N}(\bm z|\bm 0, [-\ln (1-\pi_{k})\bm D_{z}^{T}\mathop{\mathrm{diag}}(\bm \omega_{k})\bm D_{z}]^{-1}),
\end{equation}
where, the definition of $\bm D_z$ is the same as $\bm D_x$ in (\ref{eq:shape}); $\bm \omega_k$ can indicate the boundary of the $k$-th segmentation $\bm z_k$; and $\pi_k$ denotes the probability of a pixel belonging to the $k$-th class. Finally, the ground-truth label $\bm u \in \mathbb{R}^{d_{y} \times K}$ is modeled by a generalized Bernoulli distribution, i.e., $p(\bm u | \bm z) = \prod_{i=1}^{d_{y}} \prod_{k=1}^{K} z_{ik}^{u_{ik}}$, where the segmentation $\bm z$ determines the expected pixel-wise probability of the label.

Table \ref{tab:priors} shows the selected priors for $\bm m$, $\bm \rho$, $\bm \upsilon$, $\bm \omega$, and $\bm \pi$. Concretely, we assign a Gaussian prior to $\bm m$, Gamma priors to $\bm \rho$, $\bm \upsilon$ and $\bm \omega$, and a Beta prior to $\bm\pi$. Here, $\bm 1 $ denotes a vector with all elements to be ones, and $\bm I$ denotes an identity matrix. 
% \textit{Please refer to our supplementary material for the detailed formulas of these priors.} 
Note that we choose Beta distribution instead of Dirichlet distribution for $\bm\pi$, since the former is the conjugate prior of (\ref{eq:label}), which could greatly simplify our following formulation.
Through the aforementioned statistical modeling, BayeSeg can provide interpretable representations and promote the generalization ability on medical image segmentation.
%\emph{The details of Gaussian distribution, Gamma distribution, and Beta distribution are provided in our supplementary material}.

\begin{table*}
\caption{Summary of the final variational loss. Computational details and interpretation of each variational term in (\ref{eq:finalvarloss}).} \label{tab:losses}
\centering
% \resizebox{1\linewidth}{!}{
% \renewcommand{\arraystretch}{1.5}
    \begin{tabular}{lll}
    \toprule
     Modeling   & Terms of variational loss   & Semantic interpretation    \\\toprule
     Image  & $\mathcal{L}_{y}$ = $\frac{1}{2}||\bm y - (\bm x + \bm m)||^{2}_{\mathop{\mathrm{diag}}(\hat{\bm\mu}_{\rho})}$ & Achieves the fitting of an observed image $\bm y$ \\\toprule
     \multirow{2}{*}{Segmentation}  & $\mathcal{L}_{\hat{\mu}_{z}}$ = $\frac{1}{2}\sum_{k=1}^{K} c_k||\bm D_{z}\hat{\bm\mu}_{zk}||^{2}_{\mathop{\mathrm{diag}}(\hat{\bm\mu}_{\omega k})}$ & Forces spatially smoothed $\hat{\bm\mu}_{z}$ with detected boundary \\
     & $\mathcal{L}_{\hat{\sigma}_{z}}$ = $\frac{1}{2}\sum_{k=1}^{K} c_k\langle 2\hat{\bm\mu}_{\omega k}, \hat{\bm\sigma}_{z}^{2}\rangle - \langle\bm 1, \ln(\hat{\bm\sigma}_{z}^{2})\rangle$ & Prevents $q(\bm z)$ from degrading to a one-point distribution \\\toprule
    \multirow{2}{*}{Shape}  &
    $\mathcal{L}_{\hat{\mu}_{x}}$ = $\frac{1}{2}\sum_{k=1}^{K}||\bm D_{x}\hat{\bm\mu}_{x}||^{2}_{\mathop{\mathrm{diag}}(\hat{\bm\mu}_{zk}\hat{\bm\mu}_{\upsilon})}$ & Ensures spatially correlated $\hat{\bm\mu}_{x}$ with preserved edges\\
    &
    $\mathcal{L}_{\hat{\sigma}_{x}}$ = $\frac{1}{2}\sum_{k=1}^{K} \langle 2\hat{\bm\mu}_{zk}\hat{\bm\mu}_{\upsilon},\hat{\sigma}_{x}^{2}\rangle - \langle\bm 1, \ln(\hat{\bm\sigma}_{x}^{2})\rangle$ & Prevents $q(\bm x)$ from degrading to a one-point distribution \\\toprule
    \multirow{2}{*}{Appearance}  &
    $\mathcal{L}_{\hat{\mu}_{m}}$ = $\frac{1}{2}\sigma_m^{(0)}||\hat{\bm\mu}_{m}||^{2}_{2}$ & Constraints the energy of $\hat{\bm\mu}_{m}$ \\
    &
    $\mathcal{L}_{\hat{\sigma}_{m}}$ = $\frac{1}{2}[\langle\sigma_m^{(0)}\bm 1, \hat{\bm\sigma}_{m}^{2}\rangle - \langle\bm 1, \ln(\hat{\bm\sigma}_{m}^{2})]$ & Prevents $q(\bm m)$ from degrading to a one-point distribution  \\\bottomrule
    \end{tabular}
% }
\end{table*}

\subsection{Variational inference of image and label}
This section presents a variational method of inferring shape, appearance and segmentation given an image $\bm y$ and its corresponding ground-truth label $\bm u$. 
Let  $\bm\psi=\{\bm m, \bm\rho, \bm x, \bm\upsilon, \bm z, \bm\omega, \bm\pi\}$ denote the set of all variables to infer, then our aim is to estimate the posterior distribution $p(\bm\psi|\bm y, \bm u) \propto p(\bm y, \bm u | \bm\psi)p(\bm\psi)$ by maximum a \textit{posteriori} (MAP) estimation. Since direct computation is intractable, we use the variational Bayesian (VB) approach \citep{back/Blei/2017} to solve the problem. Concretely, we approximate the posterior distribution $p(\bm\psi|\bm y, \bm u)$ via a variational distribution $q(\bm\psi)$. To tackle the difficulty, the variables in $\bm\psi$ are generally assumed to be independent, namely,
\begin{equation}\label{eq:variational}
    q(\bm\psi) = q(\bm m)q(\bm\rho)q(\bm x)q(\bm\upsilon)q(\bm z)q(\bm\omega)q(\bm\pi).
\end{equation}
Since we assigned the conjugate priors to $\bm m, \bm\rho, \bm\upsilon, \bm\omega, \bm\pi$, their variational posteriors would have the same forms as given priors, as shown in Table \ref{tab:priors}. Similarly, the variational posteriors of $\bm x $ and $ \bm z$ could be successively expressed as follows,
\begin{gather}
    q(\bm x) = \mathcal{N}(\bm x | \hat{\bm \mu}_{x}, \mathop{\mathrm{diag}}(\hat{\bm\sigma}_{x}^{2})) \\
    q(\bm z) = \prod_{k=1}^{K}\mathcal{N}(\bm z| \hat{\bm \mu}_{zk}, \mathop{\mathrm{diag}}(\hat{\bm\sigma}_{zk}^{2})).
\end{gather}
To further compute the parameters of these variational posteriors, we minimize the KL divergence between $q(\bm\psi)$ and $p(\bm\psi|\bm y, \bm u)$, which results in,
\begin{align}\label{eq:KL}
&\mathop{\arg\!\min}_{q(\bm\psi)} KL(q(\bm\psi)||p(\bm\psi|\bm y, \bm u)) \nonumber \\ = &\mathop{\arg\!\min}_{q(\bm\psi)} KL(q(\bm\psi)||p(\bm\psi)) - \mathbb{E}[\ln p(\bm y, \bm u| \bm\psi)]. 
\end{align}
%\emph{The details of further unfolding the variational loss are provided in our supplementary material}. 
Moreover, we covert it to the following problem by reparameterization, 
\begin{equation} \label{eq:reparameterization}
    \mathop{\arg\!\min}_{q(\bm\psi)} KL(q(\bm\psi)||p(\bm\psi)) -\mathbb{E}_{q(\bm\rho)}[\ln p(\bm y|\bm x, \bm m, \bm\rho)] -\mathbb{E}_{q(\bm z)}[\ln p(\bm u|\bm z)].
\end{equation}
Note that minimizing the third term of (\ref{eq:reparameterization}) induces the cross-entropy loss $\mathcal L_{ce}$ between the predicted segmentation $\bm z$ and the provided manual ground-truth label $\bm u$. Next, we focus on the minimization of our variational loss $\mathcal L_{var}$,
\begin{equation}\label{eq:varloss}
  \mathop{\arg\!\min}_{q(\bm\psi)} \mathcal L_{var} = KL(q(\bm\psi)||p(\bm\psi)) - \mathbb{E}_{q(\bm\rho)}[\ln p(\bm y| \bm\psi)].
\end{equation}
\subsection{Unfolding of the variational loss}
To further understand the variational loss, we unfold it into several terms with semantic interpretation. According to the probabilistic graphical model, the objective function in (\ref{eq:varloss}) can be expressed as,
\begin{align}\label{eq:decomposion}
    &KL(q(\bm\psi)||p(\bm\psi)) -\mathbb{E}_{\bm\rho}[\ln p(\bm y|\bm x, \bm m, \bm\rho)] \nonumber \\ 
    =\: &KL(q(\bm x)q(\bm\upsilon)q(\bm z)q(\bm\omega)q(\bm\pi)||p(\bm x| \bm z, \bm\upsilon)p(\bm\upsilon)p(\bm z|\bm\omega, \bm\pi)p(\bm\omega)p(\bm\pi)) \nonumber \\
    & + KL(q(\bm m)||p(\bm m)) + KL(q(\bm\rho)||p(\bm\rho)) -\mathbb{E}_{\bm\rho}[\ln p(\bm y|\bm x, \bm m, \bm\rho)].
\end{align}

Firstly, we successively minimize (\ref{eq:decomposion}) over $q(\bm \upsilon)$, $q(\bm \omega)$, $q(\bm \pi)$, $q(\bm \rho)$ to induce the explicit formulas of computing the parameters of these posterior distributions. 
For $q(\bm\upsilon)$,  (\ref{eq:decomposion}) is minimized when
\begin{equation}
    \ln q(\bm\omega) = \expect{z,\pi}{\ln p(\bm z|\bm\omega,\bm\pi)} + \ln p(\bm\omega) + const.
\end{equation}
Solving the above equation, we have 
\begin{equation}\label{eq:upsilon}
 \hat{\bm\mu}_{\upsilon} =\frac{2\bm\gamma_{\upsilon}^{(0)} + K}{\sum_{k=1}^{K}\hat{\bm\mu}_{zk}[(\bm D_{x}\hat{\bm\mu}_{x})^{2} + 2\hat{\bm\sigma}_{x}] + 2\bm\phi_{\upsilon}^{(0)}}.
\end{equation}
Similarly, minimizing (\ref{eq:decomposion}) over $q(\bm \omega)$ leads to the formula as follows,
\begin{equation}
\hat{\bm\mu}_{\omega k} =\frac{2\bm\gamma_{\omega k}^{(0)} + 1}{c_k[(\bm D_{z}\hat{\bm\mu}_{zk})^{2} + 2{\bm\sigma}_{zk}^{2}] + 2\phi_{\omega k}^{(0)}},
\end{equation}
where, $ c_k = \Psi(\hat{\alpha}_{\pi k}+\hat{\beta}_{\pi k}) - \Psi(\hat{\beta}_{\pi k}) $, and $ \Psi(\cdot) $ denotes the Digamma function. Minimizing (\ref{eq:decomposion}) over $q(\bm \pi)$ results in the formula of computing $\hat{\alpha}_{\pi k}$ and $\hat{\beta}_{\pi k}$,
\begin{equation}
    \left\{
        \begin{aligned}
            &\hat{\alpha}_{\pi k} = \alpha_{\pi k}^{(0)} + \frac{d_{y}}{2}, \\
            &\hat{\beta}_{\pi k} = \frac{1}{2}\textstyle\sum_{i=1}^{d_y} \hat{\bm\mu}_{\omega ki} [(\bm D_{z}\hat{\bm\mu}_{zk})_{i}^{2} + 2{\bm\sigma}_{zki}^{2}] + \beta_{\pi k}^{(0)}.
        \end{aligned}
    \right.
\end{equation}
Finally, minimizing (\ref{eq:decomposion}) over $q(\bm \rho)$ yields the computation formula of $\hat{\bm\mu}_{\rho}$, namely,
\begin{equation}
    \hat{\bm\mu}_{\rho} = \frac{2\gamma_{\rho}^{(0)}+1}{[\bm y - (\bm x + \bm m)]^{2} + 2\phi_{\rho}^{(0)}}.
\end{equation}

With the computed parameters of $q(\bm \upsilon)$, $q(\bm \omega)$, $q(\bm \pi)$, $q(\bm \rho)$, we are able to further infer other posterior distributions. That is, the final variational loss regularizing the appearance, shape, and segmentation of an image is summarized as
\begin{equation}\label{eq:finalvarloss}
\mathcal L_{var} = \mathcal L_{y} + \mathcal L_{\hat{\mu}_{z}} + \mathcal L_{\hat{\sigma}_{z}}+ \mathcal L_{\hat{\mu}_{x}} + \mathcal L_{\hat{\sigma}_{x}}+ \mathcal L_{\hat{\mu}_{m}} + \mathcal L_{\hat{\sigma}_{m}},
\end{equation}
where each variational term is summarized and interpreted in Table \ref{tab:losses}. Note that this variational loss works as a whole without manually balancing its terms and minimizing (\ref{eq:finalvarloss}) is equivalent to minimizing (\ref{eq:varloss}). We unfold (\ref{eq:varloss}) for implementation and better semantic interpretation. 
% \textit{Please refer to our supplementary material for the detailed derivation of the variational loss.}

\begin{table*}[!t]
\caption{Summary of the tasks and the corresponding domains. Note that the source domains in the experiments are denoted by \dag.} \label{tab2:tasks}
\centering
    {\footnotesize
    % \resizebox{1\linewidth}{!}{
    \begin{tabular}{lllll}
        \toprule
        \multirow{2}{*}{Tasks} &
          \multirow{2}{*}{Structures} &
          \multirow{2}{*}{Original Datasets} &
          \multirow{2}{*}{Different Domains (sequence, site, modality)} &
          \multirow{2}{*}{\makecell{No. of\\ Cases}} \\
         &          &                                                                      &                                              &                    \\\toprule
        \multirow{6}{*}{Prostate} &
          \multirow{6}{*}{Prostate} &
          \multirow{2}{*}{NCI-ISBI 2013~\citep{dataset/NCI-ISBI}} &
          Radboud University Nijmegen Medical Center (RUNMC)\dag &
          30 \\
         &          &                                                                      & Boston Medical Center (BMC)                  & 30                  \\ \cline{3-5} 
         &          & I2CVB~\citep{dataset/I2CVB}                                           & Universit\'{e} de Bourgogn (UDB)             & 19                  \\ \cline{3-5} 
         &          & \multirow{3}{*}{PROMISE12~\citep{dataset/PROMISE12}}                  & University College London (UCL)              & 13                  \\
         &          &                                                                      & Beth Israel Deaconess Medical Center (BIDMC) & 12                  \\
         &          &                                                                      & Haukeland University Hospital (HK)           & 12             \\
         \bottomrule
        \multirow{6}{*}{Cardiac} &
          \multirow{6}{*}{\makecell[l]{L-Ventricle\\R-Ventricle\\Myocardium}} &
          \multirow{2}{*}{MSCMR~\citep{dataset/MSCMR/2016,dataset/MSCMR/2019}} &
          Late Gadolinium Enhanced (LGE)\dag &
          45 \\
         &          &                                                                      & T2-weighted (T2)                             & 15                  \\ \cline{3-5} 
         &          & EMIDEC~\citep{dataset/EMIDEC/2020}                                    & LGE                                          & 50                  \\ \cline{3-5} 
         &          & ACDC~\citep{dataset/ACDC/2018}                                        &Balanced Steady-state Free Precession (bSSFP)          & 50                  \\ \cline{3-5} 
         &          & MMWHS~\citep{dataset/MMWHS/2013, dataset/MMWHS/2016}& bSSFP                        & 47                  \\ \cline{3-5}
         &          & CASDC 2013~\citep{dataset/CT_RCEEAF/2013}                                                                      & CT                     & 48                 \\\bottomrule
        % \multirow{4}{*}{Abdominal} &
        %   Liver &
        %   \multirow{2}{*}{ISBI 2019 CHAOS~\citep{dataset/CHAOS}} &
        %   \multirow{2}{*}{T2 Spectral Presaturation Inversion Recovery (T2-SPIR)\dag} &
        %   \multirow{2}{*}{20} \\
        %  & L-Kidney &                                                                      &                                              &                     \\ \cline{3-5} 
        %  & R-Kidney & \multirow{2}{*}{BTVC Abdomen~\citep{dataset/BTCV}}                    & \multirow{2}{*}{CT}                          & \multirow{2}{*}{30} \\
        %  & Spleen   &                                                                      &                                              &                    \\\toprule
        \end{tabular}
        }
\end{table*}

\subsection{Neural networks and training strategy}\label{sec:train}

This section shows the network architecture for achieving the variational inference and the training strategy for image segmentation. 
\emph{At the decomposition stage}, we adopt two ResNets \citep{method/resnet/2016} as the CNNs $f_s(\cdot)$ and $f_a(\cdot)$ in Fig. \ref{fig:framework} (b) to separately infer the variational posteriors of the shape $\bm x$ and the mean of the appearance $\bm m$, i.e., $q(\bm x)$ and $q(\bm m)$, respectively. The ResNet $f_s(\cdot)$ consists of 10 residual blocks, and each block has a structure of ''Conv + ReLU + Conv''. The output of $f_s(\cdot)$ has two channels. One is the element-wise mean of the shape $\hat{\bm\mu}_{x}$, and the other is its element-wise variance $\hat{\bm\sigma}_{x}$. 
The shape $\bm x$ in the figure denotes a random sample from $q(\bm x)$. 
The ResNet $f_a(\cdot)$ consists of 6 residual blocks, and each block has a structure of ''Conv + BN + ReLU + Conv + BN''. 
Similarly, this ResNet will output the mean and variance of the $q(\bm m)$. 
Through the sampling of $\bm m$ and computation of $\hat{\bm\mu}_{\rho}$, the appearance $\bm a$ can be randomly sampled from $\mathcal{N}(\bm m, \mathop{\mathrm{diag}}(\hat{\bm\mu}_{\rho})^{-1})$. \emph{At the segmentation stage}, the segmentation network $g(\cdot)$ is adopted to infer the variational posterior of the segmentation $\bm z$, i.e., $q(\bm z)$. 
The output of this segmentation network has $2K$ channels. The first $K$ channels denote the element-wise mean of the segmentation, and the left channels represent its element-wise variance. Note that $\bm z$ in Fig. \ref{fig:framework} (b) is a random sample from the resulting posterior distribution, and it will be taken as a stochastic segmentation for training. 
In practice, we trained BayeSeg by balancing between the cross-entropy loss $\mathcal L_{ce}$ and the variational loss in (\ref{eq:finalvarloss}), which is given by,
\begin{equation}\label{eq:totalloss}
    \min_{q(\psi)} \mathcal L_{ce} + \lambda \mathcal L_{var},
\end{equation}
where, the balancing weight $\lambda$ is set to 100 in our experiments. Besides, other hyper-parameters in Fig. \ref{fig:framework} (a) are the same as the setup in our conference paper \citep{method/bayeseg_miccai/2022}, i.e.,  $\bm  \alpha_{\pi}^{(0)}=2$ and $\bm  \beta_{\pi}^{(0)}=2$; $\bm \mu_{m}^{(0)} = \bm 0$ and $\sigma_{m}^{(0)} = 1$; The elements of $\bm \gamma_{\cdot}$ are $2$ for all related variables, while the elements of $\bm  \phi_{\rho}^{(0)}$, $\bm  \phi_{\upsilon}^{(0)}$, and $\bm  \phi_{\omega}^{(0)}$ are set to $10^{-6}$, $10^{-8}$, and $10^{-4}$, respectively. 
And the necessary parameters, e.g, $\hat{\bm\mu}_{\upsilon}$ used in $\mathcal L_{var}$ are computed as Fig. \ref{fig:framework} (b) shows.
At the test stage, we took the mean of $\bm z$ as the final segmentation to get the predicted label, since the variational posterior $q(\bm z)$ is a Gaussian distribution whose mode is its mean.

\section{Experiments}
In this section, we evaluate the domain generalization performance of our BayeSeg framework on cross-sequence, cross-site and cross-modality scenarios with two tasks, i.e. prostate segmentation in subsection \ref{prostate_segmentation} and cardiac segmentation in subsection \ref{cardiac_segmentation}. Furthermore, we conduct an ablation study in subsection \ref{subsection:4.4}.

\subsection{Datasets}
% dataset, process, metric compute, pytorch implementation, data augmentation
% split, epochs, network, training, testing

For prostate segmentation, we used T2 prostate MRI images from three public datasets including NCI-ISBI 2013~\citep{dataset/NCI-ISBI}, I2CVB~\citep{dataset/I2CVB} and PROMISE12~\citep{dataset/PROMISE12}, as shown in Table \ref{tab2:tasks}. During preprocessing, we resampled all images to a fixed spacing of 0.36458$\times$0.36458 mm and clipped out the top 0.5\% of the histograms for each image. Then, the 3D images were reformatted to 2D slices, filtering out the slices without the prostate region. The slices were center-cropped into 384$\times$384 and resized to 192$\times192$. 
Z-score normalization and random data augmentation were also employed, including affine transformation, elastic transformation and additive Gaussian noise.

For cardiac segmentation, we aim to adapt segmentation models to parse three structures including left Ventricle (LV), right Ventricle (RV) and myocardium (Myo). 
As shown in Table \ref{tab2:tasks}, we employed six domains from five datasets, i.e. MSCMR~\citep{dataset/MSCMR/2016,dataset/MSCMR/2019}, EMIDEC~\citep{dataset/EMIDEC/2020}, ACDC~\citep{dataset/ACDC/2018}, MMWHS~\citep{dataset/MMWHS/2013, dataset/MMWHS/2016} and CASDC 2013~\citep{dataset/CT_RCEEAF/2013}.
% In particular, the ground-truth RV of EMIDEC and all three structures of CASDC were annotated by ourselves.
In particular, the ground-truth of EMIDEC RV and CASDC were annotated by ourselves under the supervision of experts.
We have screened and filtered out slices to align the Z-axis resolutions among datasets. Then, all remaining slices have been resampled to a fixed pixel spacing of 1.36719 $\times$ 1.36719 mm, then cropped or padded to 212$\times$212 to focus on the ROI of the heart.
We employed data augmentation with rotation, scaling, and affine transformations to reduce over-fitting during training. With the z-score normalization, all slices were normalized to a mean of zero and a standard deviation of one. Note that CT images from CASDC have been contrast enhanced through the window of [-200,300] in Housefield values before the normalization.

\subsection{Prostate segmentation} \label{prostate_segmentation}
\begin{table*}
\caption{Performance comparison with different domain generalization methods for prostate segmentation task. Here, the $\downarrow$ denotes the performance drop of mean Dice, and the best performance in terms of the mean Dice score are highlighted in bold.} \label{tab:prostate_result}

\centering
\resizebox{\linewidth}{!}{
    \begin{tabular}{|c|c|ccccc|c|}
    \hline
    \multirow{2}{*}{Method}  & RUNMC & BMC   & BIDMC   & HK   & UCL & I2CVB             & \multirow{2}{*}{\makecell{Avg. on\\Targets}}\\
    & (Source)      & \multicolumn{5}{c|}{(Target)}      &      \\\hline
    ERM     &86.6$\pm$1.0       &73.7$\pm$12.7 ($\downarrow$12.9)     &17.7$\pm$16.1 ($\downarrow$68.9)       &68.5$\pm$13.3 ($\downarrow$18.1)       &78.9$\pm$05.2 ($\downarrow$07.7)       &76.6$\pm$15.1 ($\downarrow$10.0)        &63.1    \\
    Cutout \citep{rw/data-augmentation-cutout/2017} 
            &85.5$\pm$0.7       &72.6$\pm$14.8 ($\downarrow$12.9)          &15.6$\pm$16.1 ($\downarrow$69.9)       &69.1$\pm$11.1 ($\downarrow$16.4)       &78.2$\pm$05.1 ($\downarrow$07.3)       &79.0$\pm$05.9 ($\downarrow$06.5)         &62.9    \\
    IBN-Net \citep{ibn-net} 
            &86.8$\pm$4.6       &78.4$\pm$09.0 ($\downarrow$08.4)           &42.8$\pm$14.8 ($\downarrow$44.0)       &76.0$\pm$08.9 ($\downarrow$10.8)        &75.7$\pm$06.3 ($\downarrow$11.1)       &75.9$\pm$08.4 ($\downarrow$10.9)         &69.8    \\
    RandConv \citep{rw/data-generation-Randconv/2020} 
            &86.8$\pm$4.2       &75.3$\pm$11.7 ($\downarrow$11.5)          &16.6$\pm$17.1 ($\downarrow$70.2)       &30.8$\pm$20.9 ($\downarrow$56.0)       &80.1$\pm$04.8 ($\downarrow$06.7)       &76.8$\pm$10.4 ($\downarrow$10.0)        &55.9    \\
    DSU \citep{rw/data-feature-DSU/2022} 
            &86.9$\pm$1.9       &78.7$\pm$06.3 ($\downarrow$08.2)           &31.0$\pm$24.1 ($\downarrow$55.9)       &73.9$\pm$09.5 ($\downarrow$13.0)        &78.5$\pm$06.3 ($\downarrow$08.4)       &77.6$\pm$11.1 ($\downarrow$09.3)        &67.9     \\
    BayeSeg 
    &\textbf{87.3$\pm$3.0}       &\textbf{80.9$\pm$07.0 ($\downarrow$06.4)} & \textbf{62.9$\pm$12.0 ($\downarrow$24.4)} &\textbf{83.2$\pm$03.9 ($\downarrow$04.1)} &\textbf{81.2$\pm$05.1 ($\downarrow$06.1)} & \textbf{79.1$\pm$13.7 ($\downarrow$08.2)} & \textbf{77.5}\\\hline
    \end{tabular}
}
\end{table*}

\begin{table*}[t]
\caption{Performance comparison with different domain generalization methods for cardiac segmentation task. Here, the $\downarrow$ denotes the performance drop of mean Dice, and the best performance in terms of the mean Dice score are highlighted in bold.}  \label{tab:results-cardiac}
\centering
\resizebox{1\linewidth}{!}{
% {\small
\begin{tabular}{|c|c|ccccc|c|}
    \hline
    \multirow{2}{*}{Methods} &
      LGE of MSCMR &
      T2 of MSCMR &
      LGE of EMIDEC &
      bSSFP of ACDC &
      bSSFP of MMWHS &
      CT of CASDC &
      \multirow{2}{*}{\makecell{Avg. on\\Targets}} \\
             & (Source)             & \multicolumn{5}{c|}{(Target)}                                                          &      \\ \hline
    ERM & 
        77.3$\pm$9.2 & 
        11.8$\pm$13.9 ($\downarrow$65.5)& 
        57.1$\pm$21.8 ($\downarrow$20.2)& 
        72.4$\pm$13.3 ($\downarrow$04.9)& 
        60.8$\pm$21.4 ($\downarrow$16.5)& 
        66.6$\pm$19.5 ($\downarrow$10.7)& 
        57.7 \\
    Cutout \citep{rw/data-augmentation-cutout/2017} &
      80.0$\pm$8.5 &
      19.2$\pm$14.7 ($\downarrow$60.8)&
      60.7$\pm$23.4 ($\downarrow$19.3)&
      72.1$\pm$17.6 ($\downarrow$07.9)&
      71.5$\pm$14.9 ($\downarrow$08.5)&
      67.9$\pm$20.6 ($\downarrow$12.1)&
      58.3\\ 
    IBN-Net \citep{ibn-net} &79.6$\pm$9.2
    &10.6$\pm$10.7 ($\downarrow$69.0)
    &50.0$\pm$22.2 ($\downarrow$29.6)  
    &\textbf{73.3$\pm$12.9 ($\downarrow$06.3)} 
    &73.5$\pm$12.5 ($\downarrow$06.1)  
    &73.7$\pm$13.5 ($\downarrow$05.9)
    &56.2 \\
    RandConv \citep{rw/data-generation-Randconv/2020} &
      \textbf{83.4$\pm$7.1} &
      36.3$\pm$25.8 ($\downarrow$47.1)&
      51.5$\pm$28.3 ($\downarrow$31.9)&
      71.3$\pm$17.6 ($\downarrow$12.1)&
      65.5$\pm$19.0 ($\downarrow$17.9)&
      68.6$\pm$19.3 ($\downarrow$14.8)&
      58.6\\ 
    DSU \citep{rw/data-feature-DSU/2022} &
      78.5$\pm$7.7 &
      34.0$\pm$21.5 ($\downarrow$44.5)&
      63.7$\pm$19.7 ($\downarrow$14.8)&
      71.1$\pm$14.9 ($\downarrow$07.4)&
      71.3$\pm$17.8 ($\downarrow$07.2)&
      75.6$\pm$14.8 ($\downarrow$02.9)&
      63.1 \\ 
    %BayeSeg w.o. $\mathcal{L}_{val}$ & 80.1$\pm$8.5 & 23.0$\pm$21.1 & 57.7$\pm$26.8 & \textbf{72.6$\pm$15.3} & 73.2$\pm$15.4 & 77.4$\pm$14.3 & 64.0 \\\midrule
    BayeSeg &
      80.8$\pm$7.3 &
      \textbf{70.9$\pm$13.8 ($\downarrow$09.9)} &
      \textbf{73.9$\pm$10.5 ($\downarrow$06.9)} &
      71.3$\pm$12.5 ($\downarrow$09.5)&
      \textbf{75.4$\pm$12.9 ($\downarrow$05.4)} &
      \textbf{79.2$\pm$10.5 ($\downarrow$01.6)} &
      \textbf{75.3} \\ \hline
    \end{tabular}
}
\end{table*}

In order to study the model generalizability in cross-site scenario, we used three prostate segmentation datasets composed of six domains for training and testing as shown in Table \ref{tab2:tasks}. In the training stage, we set the data from RUNMC as the source domain and randomly split its 30 cases into three sets, 21 for training, 3 for validation, and 6 for testing. We configured the widely used EfficientNet-b2 in prostate segmentation \citep{rw/feature-disentanglement-causality/2022} as our segmentation backbone. The networks were optimized by Adam up to 1200 epochs with the learning rate initialized as $3\times10^{-4}$, and the learning rate was dropped to $3\times10^{-5}$ at the 1000th epoch. We selected the best model on the validation set for testing. During the test, we set the data from sites BMC, BIDMC, HK, UCL and I2CVB as target domains. For comparison, we chose five methods, including the standard empirical risk minimization (ERM) and other four domain generalization approaches, i.e. Cutout \citep{rw/data-augmentation-cutout/2017}, IBN-Net \citep{ibn-net}, RandConv \citep{rw/data-generation-Randconv/2020} and DSU \citep{rw/data-feature-DSU/2022}. It is worth noting that, to make a fair comparison, these methods were implemented with the same segmentation backbone and training strategies as our method. After that, we computed the Dice score (0-100) for each subject to assess the performance of these methods.

\begin{figure*}[!t]
    \centering
    \subfigure[Qualitative results on prostate segmentation.]{\includegraphics[width=0.49\linewidth]{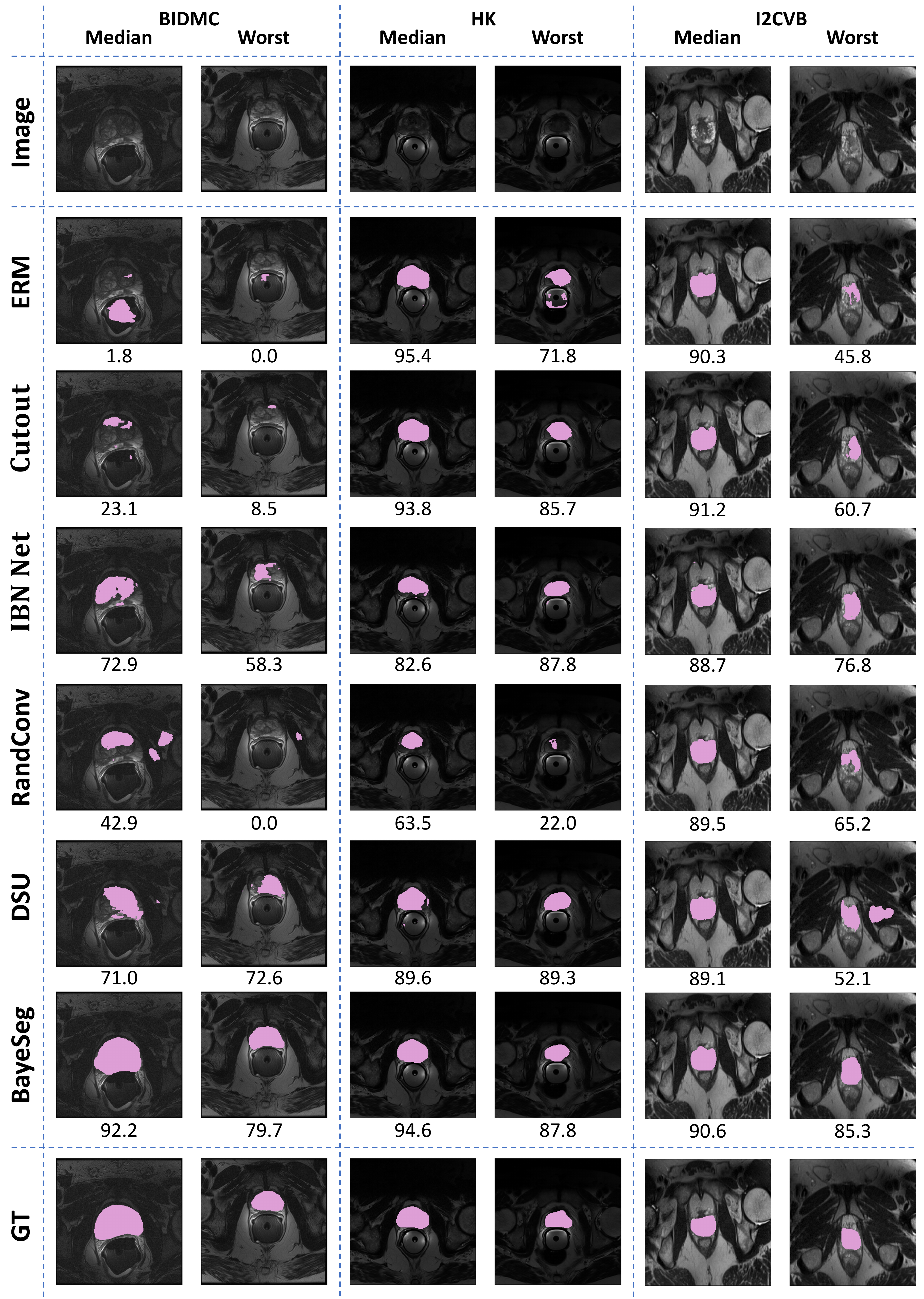}}
    \subfigure[Qualitative results on cardiac segmentation.]{\includegraphics[width=0.49\linewidth]{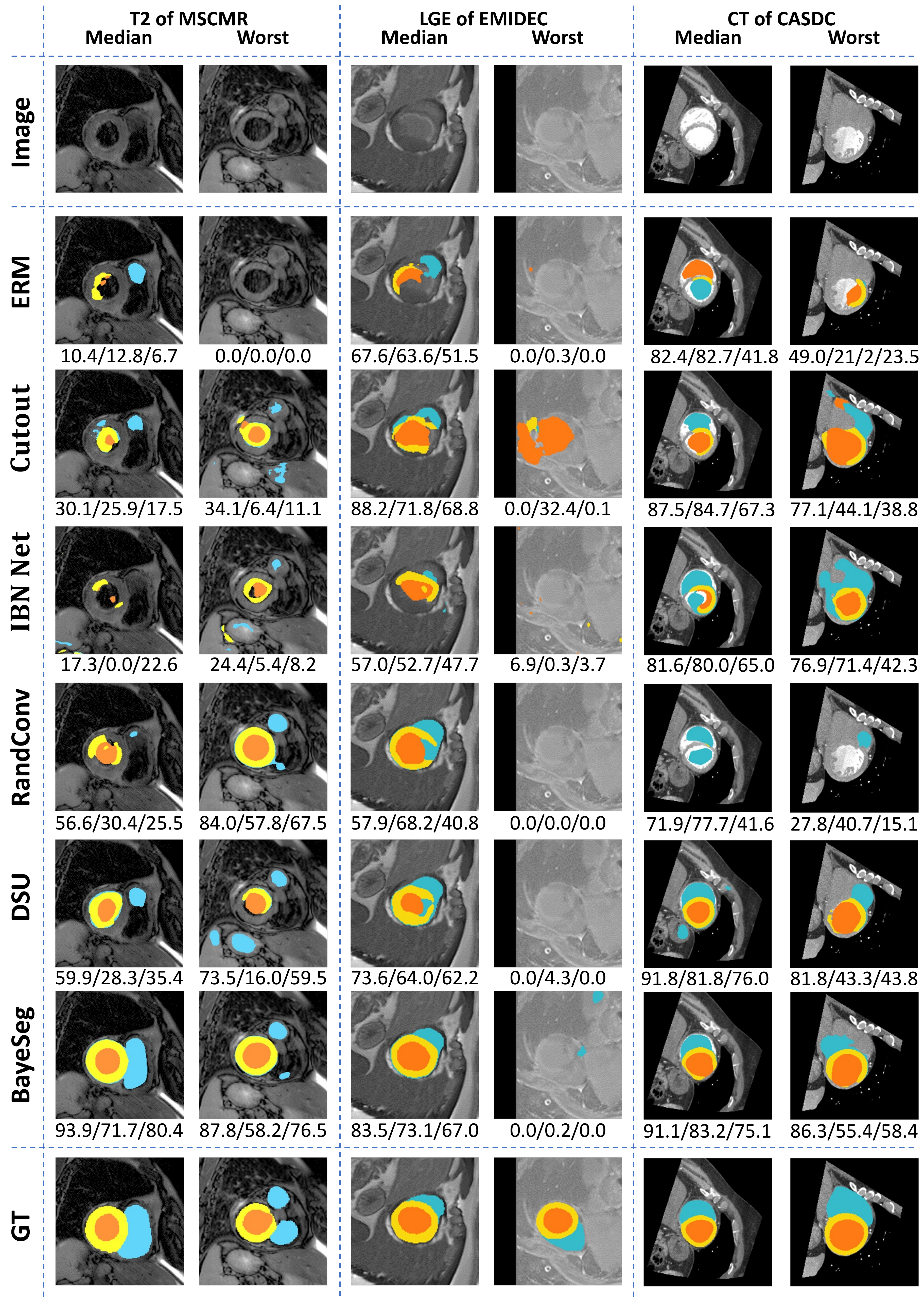}}
    \caption{Qualitative results on cardiac and prostate segmentation on different domains. For each domain, we choose the median and worst cases according to the Dice of ERM. (a) presents the qualitative results from site BIDMC, HK and I2CVB with dice below. It is worth noting that the images from T2 of MSCMR and site BIDMC are gamma transformed to change the contrast for better visualization. (b) shows the results from T2 of MSCMR, LGE of EMIDEC and CT and CASDC domains, which cover cross-sequence, cross-site and cross-modality scenarios. Dice scores are presented for each structure as LV/RV/Myo.}
    \label{fig:results}
\end{figure*}

Table \ref{tab:prostate_result} reports the mean and standard deviation of the dice score, as well as the dice drop compared with values of the source domain. Overall, BayeSeg achieves an average dice score of 77.5, surpassing the second-best method by 7.7. For the data from BIDMC with wild bias fields, four out of six methods fail to segment the prostate, since the mean and standard deviation of their dice score have close numerical values. Among the successful methods, our method exceeds the second-best method over 20, with the least performance drop of 24.4. For the data from HK with mild bias fields, among the five successful methods, BayeSeg gets a slight performance drop of 4.1, which is also the least. In other sites, all the methods achieve comparable results, with little performance gap, meaning that the domain shifts between target and source are less. We visualized the median and worst cases from the three sites shown in Fig. \ref{fig:results} (a). It can be observed that only two methods (DSU and BayeSeg) succeed in all cases. But, in the worst case of I2CVB, DSU generates a disconnected segmentation, which violates the anatomy of prostates. Instead, BayeSeg keeps connected segmentation results for all cases, benefiting from the spatial correlation modeling.

\subsection{Cardiac segmentation} \label{cardiac_segmentation}

Beyond the cross-site scenario, we further explored the model generalizability in cross-sequence and cross-modality settings, using five cardiac segmentation datasets, which consist of six domains as shown in Table \ref{tab2:tasks}. In the training stage, we set LGE of MSCMR as source domain and randomly split its 45 cases into three sets, 25 for training, 5 for validation, and 15 for test. After that, we followed the setups in our conference paper \citep{method/bayeseg_miccai/2022}, choosing the U-net \citep{method/unet/2015} as the segmentation backbone. Finally, we trained BayeSeg by Adam optimizer up to 2000 epochs with $10^{-4}$ learning rate and selected the best model on the validation set for testing. During test, we set T2 of MSCMR, LGE of EMIDEC, bSSFP of ACDC, bSSFP of MMWHS and CT of CASDC as target domains. For comparison, we adopted the same comparative methods and evaluation protocols as prostate segmentation in Section \ref{prostate_segmentation}.

To evaluate model generalizability quantitatively, we report the mean, standard deviation and the drop of dice score in Table \ref{tab:results-cardiac}. The table shows that performance of BayeSeg on target domains is the closest to that of the source domain, with only 5.5 dice drop. Particularly, for the cross-sequence scenario (T2 of MSCMR), all the counterpart methods obtain dice scores lower than 36.3, of which the least performance drop is higher than 44.5. In contrast, BayeSeg demonstrates the best performance in the T2 domain of MSCMR, surpassing the second-best method by an impressive margin of 34.6, with a small performance drop of 9.9. For cross-site (LGE of EMIDEC) and cross-modality (CT of CASDC) scenarios, BayeSeg outperforms the second-best method by 10.2 and 3.6 respectively. For qualitative evaluation, we visualized typical cases from three domains representing cross-sequence, cross-site and cross-modality respectively in Fig. \ref{fig:results} (b). The figure shows that, in five out of six cases, the proposed BayeSeg delivers comparable or superior performance. However, all the methods fail in the worst case of EMIDEC LGE, demonstrating that our method still has room for improvement.

%Due to the benefit of stochastic mapping, Baseline had better generalization ability than ERM, as shown in Table 3 and 4. Concretely, BayeSeg first maps an image to stochastic samples from the distribution of shape, and then maps them to stochastic samples from the distribution of segmentation. Theoretically, this nature makes BayeSeg see infinite samples with respect to shape for a given image, and thus has the potential of improving model generalizability. Since Baseline shares the same network architecture with BayeSeg, it also benefited from the stochastic mapping.    

\subsection{Ablation study on basic components}\label{subsection:4.4}

\begin{table*}[!t] 
\caption{Effects of stochastic mapping and variational loss on BayeSeg for cardiac segmentation. Here, the best performance in terms of the mean Dice score are highlighted in bold.}\label{tab5:basicablation}
\centering
 \resizebox{1\linewidth}{!}{
\begin{tabular}{|c|cc|c|ccccc|}
%\multicolumn{2}{|c|}{BayeSeg} & \multicolumn{6}{c|}{}\\
\hline
\multirow{2}{*}{Model}&\multirow{2}{*}{Stochastic mapping} & \multirow{2}{*}{Variational loss} &LGE of MSCMR & T2 of MSCMR  & LGE of EMIDEC & bSSFP of ACDC & bSSFP of MMWHS & CT of CASDC\\
 &&&(Source) & \multicolumn{5}{c|}{(Target)}\\
 \hline
\#1 &N & N  &79.1$\pm$9.4
 &9.83$\pm$10.3
 &57.6$\pm$24.5
 &72.7$\pm$14.2
 &\textbf{73.7$\pm$14.2}
 &75.7$\pm$14.5
\\ 
\#2 & Y & N & 79.0$\pm$9.3 
& 9.5$\pm$11.8 
& 55.3$\pm$26.4 
& 72.4$\pm$14.5
& 71.5$\pm$16.3 
& 76.7$\pm$14.3
\\ 
\#3 (proposed) &Y&Y&\textbf{80.9$\pm$6.8}
 &\textbf{63.1$\pm$17.6} 
 &\textbf{67.0$\pm$16.4}
 &\textbf{74.1$\pm$12.0}
 &73.5$\pm$14.1
 &\textbf{79.4$\pm$10.2}
\\ 
\hline
\end{tabular}}
\end{table*}

In order to explore the effectiveness of stochastic mapping and variational loss in BayeSeg, we conducted ablation studies on cardiac segmentation task in this subsection. Particularly, we evaluated the generalizability of different BayeSeg settings. To reduce the impact of randomness, we performed testing every 100 epochs when the training process exceeds 1000 epochs and obtained the final result by averaging 10 evaluations. This rigorous approach allows us to thoroughly analyze the impact of individual components on the overall generalizability of BayeSeg.

As Table \ref{tab5:basicablation} presents, the variational loss is the key factor to BayeSeg's generalizability. With the help of variational loss, the shape is able to model the domain-invariant structure information of interested organs, which enhances the segmentation generalizability on unseen domains.
Additionally, one can see that introducing stochastic mapping alone can not bring remarkable performance improvement. However, it is worth noting that using stochastic mapping is a prerequisite for applying the variational loss. 
Under the guidance of variational loss, 
stochastic mapping makes BayeSeg see infinite samples with respect to shape for a given image, and thus also helps improve performance on unseen domains.   
%Concretely, BayeSeg first maps an image to stochastic samples from the distribution of shape, and then maps them to stochastic samples from the distribution of segmentation. Theoretically, this nature makes BayeSeg see infinite samples with respect to shape for a given image, and thus has the potential of improving performance on unseen domains. 

\section{Study of interpretability and generalizability\label{section:5}}
\begin{figure*}[t]
    \centering
    \includegraphics[width=0.9\linewidth]{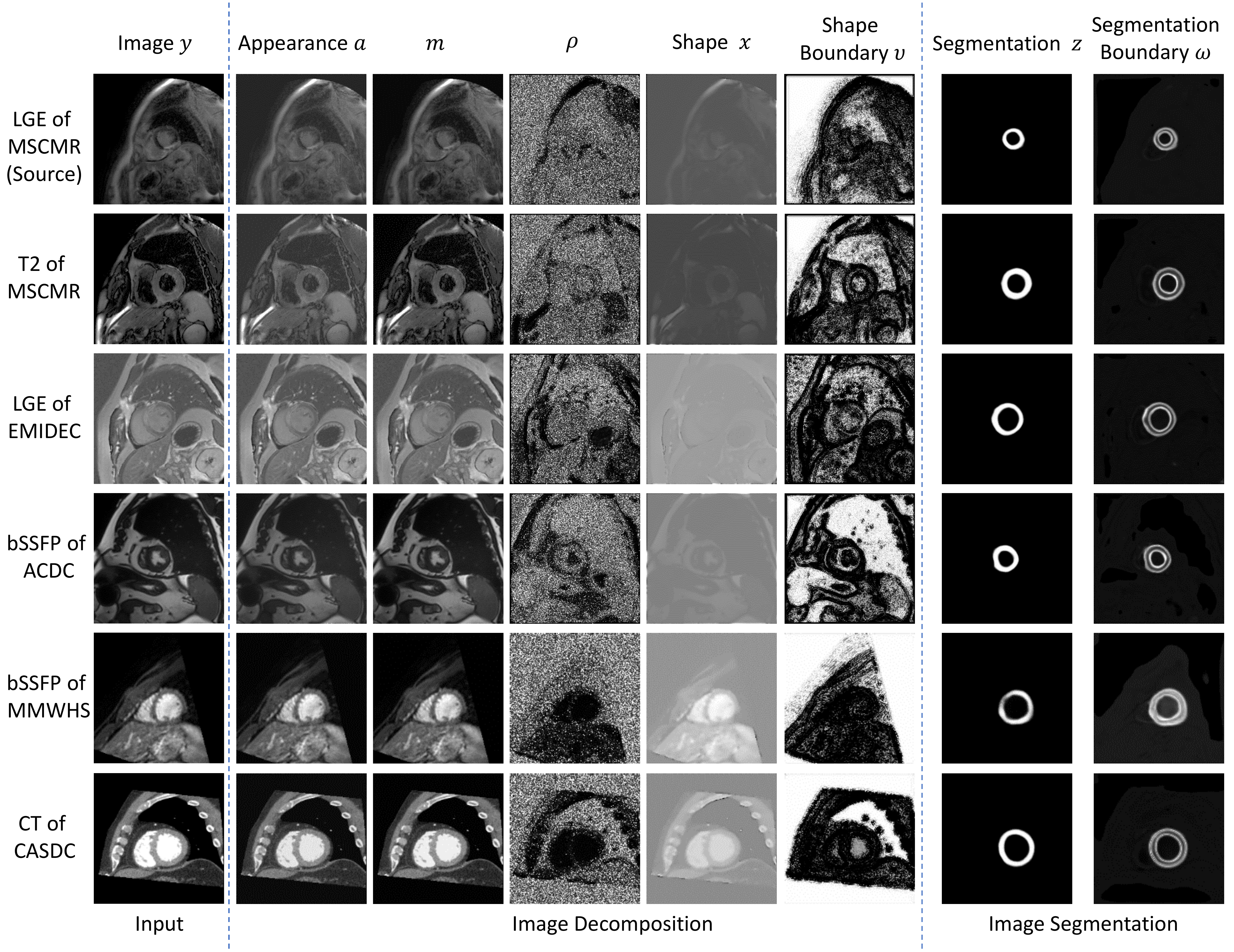}\\[-2.5ex]
    \caption{Visualization of posteriors inferred by BayeSeg. Here, the segmentation $\bm z$ and the segmentation boundary $\bm\omega$ of myocardium are visualized. Note that the visualized appearance $\bm a$, shape $\bm x$ and segmentation $\bm z$ are randomly sampled, while other variables are the expectations of posteriors.}
    \label{fig:img-posteriors}
\end{figure*}

In this section, we investigate the interpretability and generalizability of BayeSeg. 
In subsection \ref{subsection:5.1}, we explain the posteriors of variables inferred by BayeSeg from the overall.
Then, in subsection \ref{subsection:5.2}, we further explore the influence of changing priors on the top layer variables $\bm \rho$, $\bm \upsilon$ and $\bm \omega$, to show the benefits of statistical prior modeling in BayeSeg.
Finally, in subsection \ref{subsection:5.3}, we further test the influence on generalization ability from restraining variables in the second layer, i.e., $\bm z$ and $\bm a$ to present the benefits of joint modeling of images and labels. Note that the evaluation strategy in this section is the same as that in subsection \ref{subsection:4.4}.

\subsection{Interpretation of statistical representations}\label{subsection:5.1}
In this subsection, we present the interpretability of BayeSeg through the visualization of posteriors. According to the interpretability taxonomy from \cite{back/interpretability_survey/2021}, BayeSeg's interpretability can be presented from three dimensions. 
Firstly, BayeSeg achieves active interpretability by enforcing extracted variables to adhere to specific variational posterior distributions, which are predetermined through a statistical model designed prior to the training process. The induced variational loss ensures that the resulting representations are both informative and interpretable. Additionally, by modeling the statistical representations as stochastic variables, BayeSeg offers global interpretability over the distribution of input images, rather than just individual samples. Furthermore, from the perspective of explanation type, as illustrated in Fig. \ref{fig:img-posteriors}, BayeSeg provides semantic interpretation of the extracted representations, thereby enhancing the comprehensibility and interpretability of the model.

\begin{table*}[!t]
\caption{Quantitative average Dice score of setting different Gamma prior $\mathcal{G}(\phi^{(0)}_{\rho},\gamma^{(0)}_{\rho})$ for $\bm \rho$. Here, ``one-dim'' in model \#2 denotes setting $\bm \rho$ as a scalar.}\label{tab5:rho}
\centering
 \resizebox{1\linewidth}{!}{
\begin{tabular}{|c|c|c|ccccc|}
\hline
\multirow{2}{*}{Model}&\multirow{2}{*}{Prior of $\bm \rho$} &LGE of MSCMR & T2 of MSCMR  & LGE of EMIDEC & bSSFP of ACDC & bSSFP of MMWHS & CT of CASDC\\
 &&(Source) & \multicolumn{5}{c|}{(Target)}\\
 \hline
 \#1  (proposed)
 &$\mathcal{G}(10^{-6},2)$ &\textbf{80.9$\pm$06.8}
 &\textbf{63.1$\pm$17.6} 
 &\textbf{67.0$\pm$16.4}
 &\textbf{74.1$\pm$12.0}
 &73.5$\pm$14.1
 &79.4$\pm$10.2
\\ 
 \#2
 &$\mathcal{G}(10^{-6},2)$ (one-dim)&77.5$\pm$09.9
 &32.8$\pm$17.0
 &57.3$\pm$21.5
 &68.6$\pm$13.6
 &61.6$\pm$14.8
 &60.0$\pm$15.0
\\
 \#3
 &$\mathcal{G}(10^{-4},2)$&77.3$\pm$10.7
 &32.1$\pm$13.8
 &52.8$\pm$20.8
 &71.9$\pm$12.7
 &61.9$\pm$15.4
 &64.2$\pm$13.8
\\
\#4
 &$\mathcal{G}(10^{-8},2)$&79.0$\pm$08.6
 &16.5$\pm$14.4
 &63.8$\pm$19.0
 &73.5$\pm$12.5
 &73.2$\pm$14.8
&\textbf{78.3$\pm$11.8}
\\
 \#5
 &$\mathcal{G}(10^{-6},0.2)$&69.4$\pm$09.8
 &28.3$\pm$12.4
 &47.7$\pm$21.4
 &63.0$\pm$12.9
 &51.6$\pm$13.9
 &55.3$\pm$12.2
 \\
 \#6
 &$\mathcal{G}(10^{-6},20)$&79.8$\pm$08.3
 &10.2$\pm$10.2
 &60.0$\pm$22.7
 &74.0$\pm$14.1
 &\textbf{73.7$\pm$14.4}
 &70.7$\pm$16.5 
 \\
%$\mathcal{G}(10^{-8},2)$& & & & & & & \\
\hline
\end{tabular}}
\end{table*}
\begin{figure*}[t]
    \centering
    \includegraphics[width=0.85\linewidth]{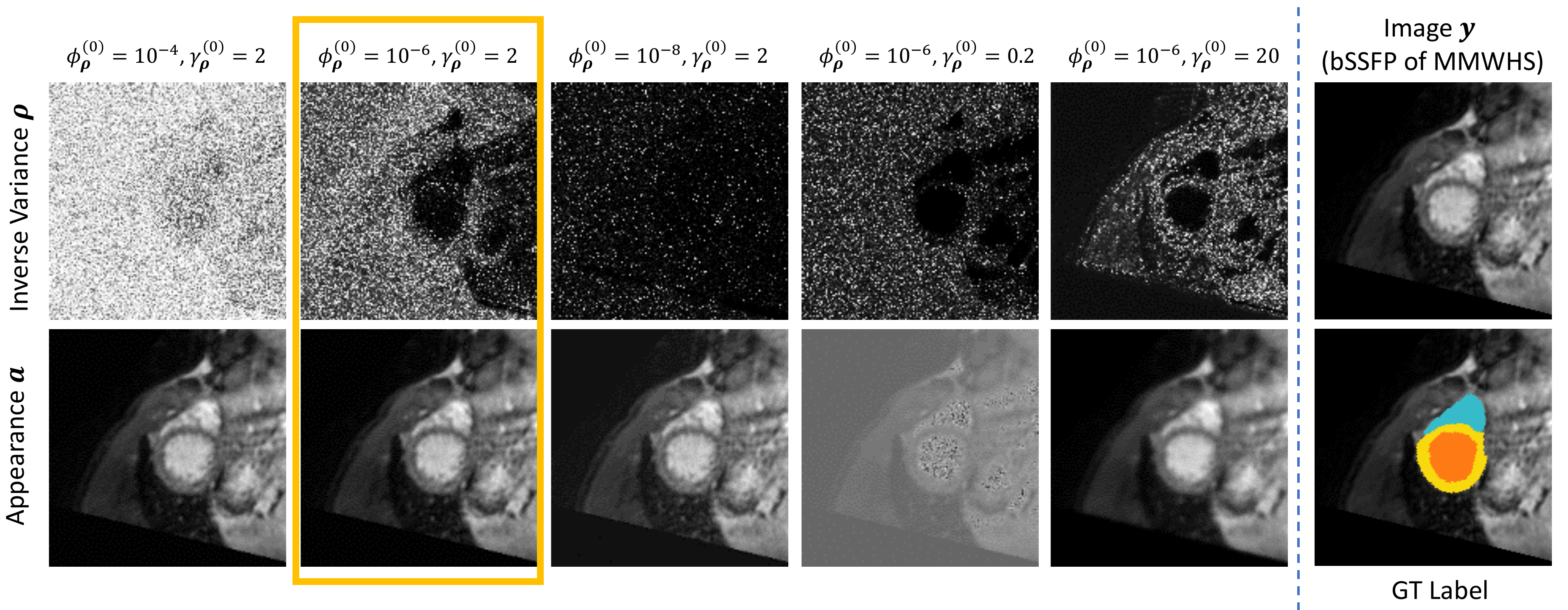}\\[-3ex]
    \caption{Visualization of posteriors of $\bm \rho$ (upper) and $\bm a$  (lower) under different priors of $\bm \rho$. The results of the proposed prior are highlighted by box.}
    \label{fig:rho}
\end{figure*}
% Firstly, BayeSeg achieves active interpretability, forcing the extracted variables to follow specific variational posterior distributions in the form of regularization terms, which are induced from the statistical model designed before training process.
% Secondly, since we model the statistical representations as stochastic variables, BayeSeg has the capability to provide global interpretability over the distribution of input images, not limited to individual samples. 
% Finally, from the perspective of explanation type, as Fig. \ref{fig:img-posteriors} shows, BayeSeg brings semantic interpretation on extracted representations. 
One can see that, at the decomposition stage, the input image is decomposed into its appearance and shape. The appearance $\bm a$ is modeled as a spatial-variant Gaussian distribution with the mean $\bm m$ and the inverse variance $\bm \rho$. It is an approximation of the image. The shape $\bm x$ is modeled to be spatial-correlated to capture the structure information. Therefore, we assign the shape boundary $\bm \upsilon$ to detect its edges. As shown in the third row of Fig. \ref{fig:img-posteriors}, the $\bm \upsilon$ successfully detects the region of the left ventricle, which is not obvious to recognize in the raw input image.
At the segmentation stage, we choose to segment the shape, since it is more likely to be sequence-independent, site-independent, and even modality-independent. To achieve better segmentation around the boundary of some objects, such as myocardiums, we assign the $\omega$ to detect the segmentation boundary. This variable can effectively indicate the inner and outer boundaries of myocardium. 

Moreover, based on the configuration of the probabilistic graphical model, we interpret the hypothesis of BayeSeg's generalization ability from two directions for further exploration, i.e., the explicit statistical prior modeling, and the joint modeling of image and label statistics which enhances the flexibility of BayeSeg.
%On the one hand, the explicit statistical prior modeling ensures the stability of BayeSeg. It forces variables like shape $\bm x$ to follow a specific variational posterior distribution, restricting their distribution shift on unseen data. 
%On the other hand, the stochastic nature of hierarchical bayes modeling enhances the flexibility of BayeSeg. The stochastic variables which models the domain-related appearance, like appearance $\bm \rho$, allocates more room for adjustment to different domains.
 Further ablation study results and corresponding discussions are presented in the following subsections \ref{subsection:5.2} and \ref{subsection:5.3}.

\subsection{Generalizability of statistical prior modeling} \label{subsection:5.2}
To further investigate the effect of statistical prior modeling in BayeSeg, we conducted ablation studies on three pairs of hyper-parameters in Gamma priors, i.e., $\bm \phi_{\omega}^{(0)}$, $\bm \gamma_{\omega}^{(0)}$,  $\bm \phi_{\upsilon}^{(0)}$, $\bm \gamma_{\upsilon}^{(0)}$, $\bm \phi_{\rho}^{(0)}$, $\bm \gamma_{\rho}^{(0)}$, which respectively influence the variational posterior distribution of three important variables $\bm \omega$, $\bm \upsilon$ and $\bm \rho$ in the top layer of the probabilistic graphical model.  

\subsubsection{Study of the $\rho$ prior}

$\bm \rho$ serves as the inverse variance of the appearance $\bm a$ in Gaussian distribution. The large $\bm \rho$ indicates the smooth region of appearance, while the small $\bm \rho$ models fine textures that would vary greatly.
Due to the decomposition $\bm y =\bm x +\bm a$, it also indirectly affects the shape $x$. Hence, the influence of  $\bm \rho$ on model generalizability is comprehensive.  
Table \ref{tab5:rho} provides the generalization results of BayeSeg under different $\bm \rho$ priors. Here, our proposed prior achieves superior average performance on three out of five target domains, especially on the most difficult domain, i.e. T2 of MSCMR.
Intuitively, on the one hand, priors that tend to induce too large values of $\bm \rho$ overly reduce the variance of $\bm a$, which may constrain the adaptability of appearance $\bm a$ to domain-related details. On the other hand, priors that tend to result in too small values of $\bm \rho$ overly increase the variance of $\bm a$, making BayeSeg more difficult to converge in the training procedure.
Moreover, setting $\bm \rho$ as a one-dim scalar indicates the variance of appearance is spatially consistent, which restricts the ability to model image details and thus affect the generalization ability of the model. 
As it can be seen from the visualization results in Fig. \ref{fig:rho}, a proper prior for $\bm \rho$ helps BayeSeg capture the inter-structure details in appearance.

\subsubsection{Study of the $\upsilon$ prior}
\begin{table*}[!t] 
\caption{Quantitative average Dice score of setting different Gamma prior $\mathcal{G}(\phi^{(0)}_{\upsilon},\gamma^{(0)}_{\upsilon})$ for $\bm \upsilon$.}\label{tab5:upsilon}
\centering
 \resizebox{1\linewidth}{!}{
\begin{tabular}{|c|c|c|ccccc|}
\hline
\multirow{2}{*}{Model}&\multirow{2}{*}{Prior of $\bm \upsilon$}
 &LGE of MSCMR & T2 of MSCMR  & LGE of EMIDEC & bSSFP of ACDC & bSSFP of MMWHS & CT of CASDC\\
 &&(Source) & \multicolumn{5}{c|}{(Target)}\\
 \hline
 \#1  (proposed)
 &$\mathcal{G}(10^{-8},2)$ &\textbf{80.9$\pm$06.8}
 &\textbf{63.1$\pm$17.6} 
 &67.0$\pm$16.4
 &74.1$\pm$12.0
 &73.5$\pm$14.1
 &\textbf{79.4$\pm$10.2}
\\ 
 \#2
 &$\mathcal{G}(10^{-4},2)$&79.7$\pm$08.3
 &26.3$\pm$20.7
 &59.4$\pm$24.7
 &\textbf{74.7$\pm$12.0}
 &\textbf{75.6$\pm$13.9}
 &78.7$\pm$11.2
 \\
 \#3
 &$\mathcal{G}(10^{-6},2)$&79.2$\pm$07.8
 &49.5$\pm$19.8
 &\textbf{67.2$\pm$18.1}
 &72.8$\pm$13.2
 &73.9$\pm$14.7
 &76.6$\pm$12.3
\\
 \#4
 &$\mathcal{G}(10^{-10},2)$&80.3$\pm$08.0
 &31.5$\pm$17.7
 &65.6$\pm$17.3
 &72.6$\pm$13.9
 &72.3$\pm$14.7
 &77.2$\pm$12.5
\\
 \#5
 &$\mathcal{G}(10^{-8},0.2)$&79.2$\pm$09.0
 &33.9$\pm$21.6
 &66.8$\pm$17.6
 &72.4$\pm$12.8
 &72.7$\pm$15.1
 &65.3$\pm$14.1
\\
 \#6
 &$\mathcal{G}(10^{-8},20)$&78.2$\pm$10.1
 &38.4$\pm$19.9
 &60.7$\pm$20.4
 &72.0$\pm$13.1
 &64.8$\pm$15.0
 &77.5$\pm$11.9
\\
\hline
\end{tabular}}
\end{table*}
\begin{figure*}[t]
    \centering
    \includegraphics[width=0.85\linewidth]{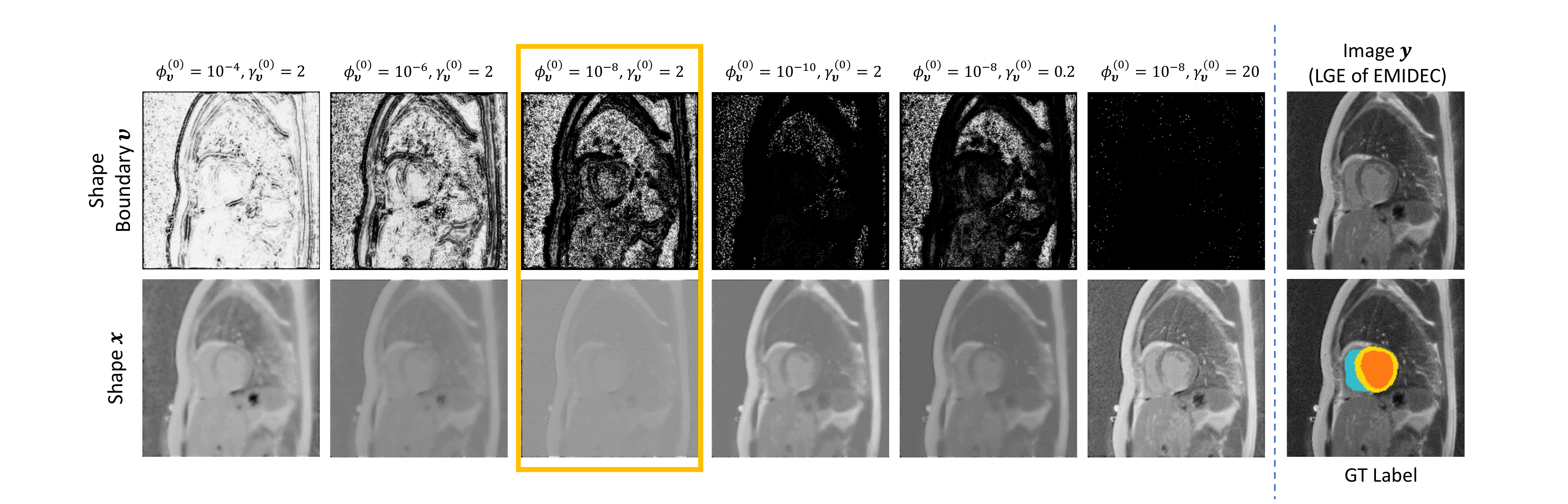}\\[-3ex]
    \caption{Visualization of posteriors of $\bm \upsilon$ (upper) and $\bm x$  (lower) under different priors of $\bm \upsilon$. The results of the proposed prior are highlighted by box.}
    \label{fig:upsilon}
\end{figure*}

To avoid the smoothness of shape $\bm x$, the line $\bm \upsilon$ is assigned to detect and enhance its edges. The large value of $\bm \upsilon$ shows the smooth areas of shape, while the small values indicate the edges. Consequently, $\bm \upsilon$ also has a strong impact on the generalization ability of BayeSeg. Fig. \ref{fig:upsilon} visualizes the posteriors under different $\bm \upsilon$ priors. One can see that setting sharper priors (e.g., $\mathcal{G}(10^{-4},2)$) results in sparser $\bm \upsilon$. It tends to neglect some important edges, regarding it as the inside of structures. Therefore, its corresponding shape is visually closer to the over-smoothed version of observation $\bm y$, which fails to capture the shape of structures. Results in Table \ref{tab5:upsilon} further show that although model with prior $\mathcal{G}(10^{-4},2)$ has achieved best average Dice on bSSFP of ACDC and bSSFP of MMWHS, it failed to generalize well on the most difficult target domain T2 of MSCMR. Contrarily, assigning flatter priors (e.g., $\mathcal{G}(10^{-10},2)$) to $\bm \upsilon$ makes its posterior more sensitive to edges, thus it will induce more noises, making it difficult to distinguish target structures from backgrounds, which also leads to inferior performances.

\subsubsection{Study of the $\omega$ prior}

\begin{figure*}[t]
    \centering
    \includegraphics[width=0.85\linewidth]{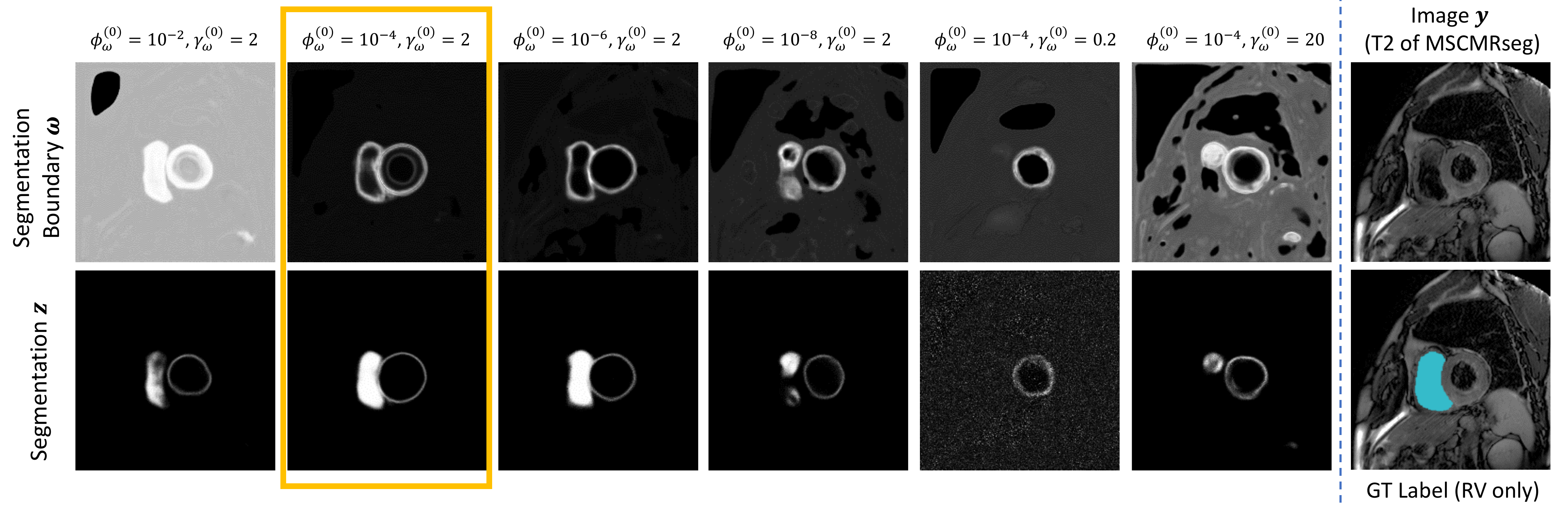}\\[-3ex]
    \caption{Comparison over the posteriors of $\bm \omega$ (upper) and $\bm z$  (lower) for the left ventricle (LV) under different priors of $\bm \omega$. The results of the proposed prior are highlighted by box.}
    \label{fig:omega}
\end{figure*}

As the boundary of segmentation $\bm z$, $\bm \omega$ regularizes the spatial relevance among pixels in $\bm z$. The large $\bm \omega_{ki}$ indicates the regions that are difficult to segment, reflecting the corresponding high uncertainty in $\bm z$.
%restricts the change of the $k$-th segmentation logits among the area surrounding position $i$, while the small $\bm \omega_{ki}$ makes logits capable for more change.
Results from Table \ref{tab5:omega} reveal the influence of changing the prior of $\omega$. One can see that the proposed setting has achieved the overall best performances. 
As Fig. \ref{fig:omega} further presents, the Gamma prior of $\bm \omega$ controls the dependence of BayeSeg's decision on predicted segmentation. For instance, setting $\phi^{(0)}_{\omega}=10^{-6}$ and $\gamma^{(0)}_{\omega}=20$ would make BayeSeg tend to highlight the surrounding background, unnecessarily taking the background into consideration, which is not favored to satisfactory generalization.

\begin{table*}[!t]
\caption{Quantitative average Dice score of setting different Gamma prior $\mathcal{G}(\phi^{(0)}_{\omega},\gamma^{(0)}_{\omega})$ for $\bm \omega$.}\label{tab5:omega}
\centering
 \resizebox{1\linewidth}{!}{
\begin{tabular}{|c|c|c|ccccc|}
\hline
\multirow{2}{*}{Model}
 &\multirow{2}{*}{Prior of $\bm \omega$}
 &LGE of MSCMR & T2 of MSCMR  & LGE of EMIDEC & bSSFP of ACDC & bSSFP of MMWHS & CT of CASDC\\
 &&(Source) & \multicolumn{5}{c|}{(Target)}\\
 \hline
  \#1 (proposed)
 &$\mathcal{G}(10^{-4},2)$  &\textbf{80.9$\pm$06.8}
 &\textbf{63.1$\pm$17.6} 
 &\textbf{67.0$\pm$16.4}
 &74.1$\pm$12.0
 &73.5$\pm$14.1
 &\textbf{79.4$\pm$10.2}
\\ 
 \#2
 &$\mathcal{G}(10^{-2},2)$&79.7$\pm$07.8
&51.4$\pm$22.9
&62.3$\pm$19.4
&73.9$\pm$12.1
&74.5$\pm$14.0
&78.7$\pm$10.7
\\
 \#3
 &$\mathcal{G}(10^{-6},2)$&79.9$\pm$07.3
&49.2$\pm$14.6
&60.5$\pm$20.9
&73.6$\pm$12.5
&\textbf{74.8$\pm$13.9}
&74.9$\pm$12.2
\\
 \#4
 &$\mathcal{G}(10^{-8},2)$&78.2$\pm$08.5
 &48.5$\pm$21.9
 &66.5$\pm$17.4
 &74.2$\pm$11.7
 &74.6$\pm$12.9
 &74.4$\pm$12.8
\\
 \#5
 &$\mathcal{G}(10^{-4},0.2)$&80.0$\pm$08.2
 &54.1$\pm$22.0
 &65.3$\pm$18.2
 &\textbf{74.4$\pm$12.6}
 &73.9$\pm$14.4
 &78.0$\pm$11.6
\\
 \#6
 &$\mathcal{G}(10^{-4},20)$&79.7$\pm$08.1
 &52.3$\pm$20.6
 &66.0$\pm$18.7
 &74.3$\pm$12.2
 &73.1$\pm$15.4
 &78.0$\pm$12.1
\\
%$\mathcal{G}(10^{-6},20)$&76.8$\pm$9.4
 %&32.7$\pm$16.2
 %&58.3$\pm$20.0
 %&71.3$\pm$13.4
% &70.8$\pm$15.2
 %&40.1$\pm$23.5
 %& \\
\hline
\end{tabular}}
\end{table*}

To sum up, statistical prior modeling plays an essential role in enhancing model generalizability. It forces variables to follow a specific variational posterior distribution, making features like shape $\bm x$ of the target domain stay close to those of the source domain.
From the perspective of target datasets, compared to easy target domains (e.g. bSSFP of MMWHS), the influence of proper priors are greater on difficult target domain (e.g. T2 of MSCMR). 
%It is potentially because the feature spaces of those easy target domains are already close enough to that of the source domain, hence the constraint proper priors brings on the posterior of shape would not benefit so much on preventing from distribution shift. 
From the perspective of variables, as one can see from quantitative results, changing priors of $\bm \rho$ and $\bm \upsilon$ has a more obvious effect than $\bm \omega$, which is consistent with their roles in statistical modeling. The shape and appearance are adaptively balanced after selecting proper $\gamma_{\rho}^{(0)}$ and $\gamma_{\upsilon}^{(0)}$, therefore, BayeSeg could properly decompose an image into the shape and appearance due to the priors with respect to $\bm \upsilon$ and $\bm \upsilon$, serving the afterward segmentation as a solid foundation. The BayeSeg framework offers a distinct advantage over traditional deep learning methods by providing greater interpretability, which enables a more nuanced understanding of how priors impact the model's performance.

\subsection{Generalizability of the joint modeling}\label{subsection:5.3}

To further explore the benefits of the joint modeling of image and label statistics in BayeSeg, we evaluated BayeSeg under two simplified PGMs. 
In one configuration, we set the appearance $\bm a$ to be deterministic, which results in the modification of both stochastic mapping and the variational loss in BayeSeg framework. On the one hand, the corresponding CNN $f_a(\cdot)$ only has one channel for output. It directly extracts the exact value of the appearance, rather than sampling from the posterior distribution. On the other hand, the $\bm m$ and $\bm \rho$ are pruned together, inducing the simplified variational loss without $\mathcal{L}_{y}$, $\mathcal{L}_{\hat{\mu}_{m}}$ and $\mathcal{L}_{\hat{\mu}_{m}}$ in the training process. 
Similarly, in another PGM configuration, the segmentation $\bm z$ is set to be deterministic, hence the network $g(\cdot)$ and variational loss are modified correspondingly.
%\emph{The details of modified variational loss in two configurations are provided in our supplementary material}.

Table \ref{tab5:config} presents the overall performance of BayeSeg under different PGM configurations. As one can see, Model $\#1$ which denotes BayeSeg with the full PGM outperforms other configurations. 
As the results of model \#2, \#3, and \#4 show, forcing $\bm a$ to be deterministic can reduce the generalization ability, especially on T2 of MSCMR, since  the variable $\bm a$ models of appearance-related image details. Similarly, from the results of model \#5, \#6, and \#7, one can see that setting $\bm z$ to be deterministic also results in inferior generalization performance due to the lack of label statistics modeling.

\begin{figure}[!t]
    \centering
    \includegraphics[width=0.95\linewidth]{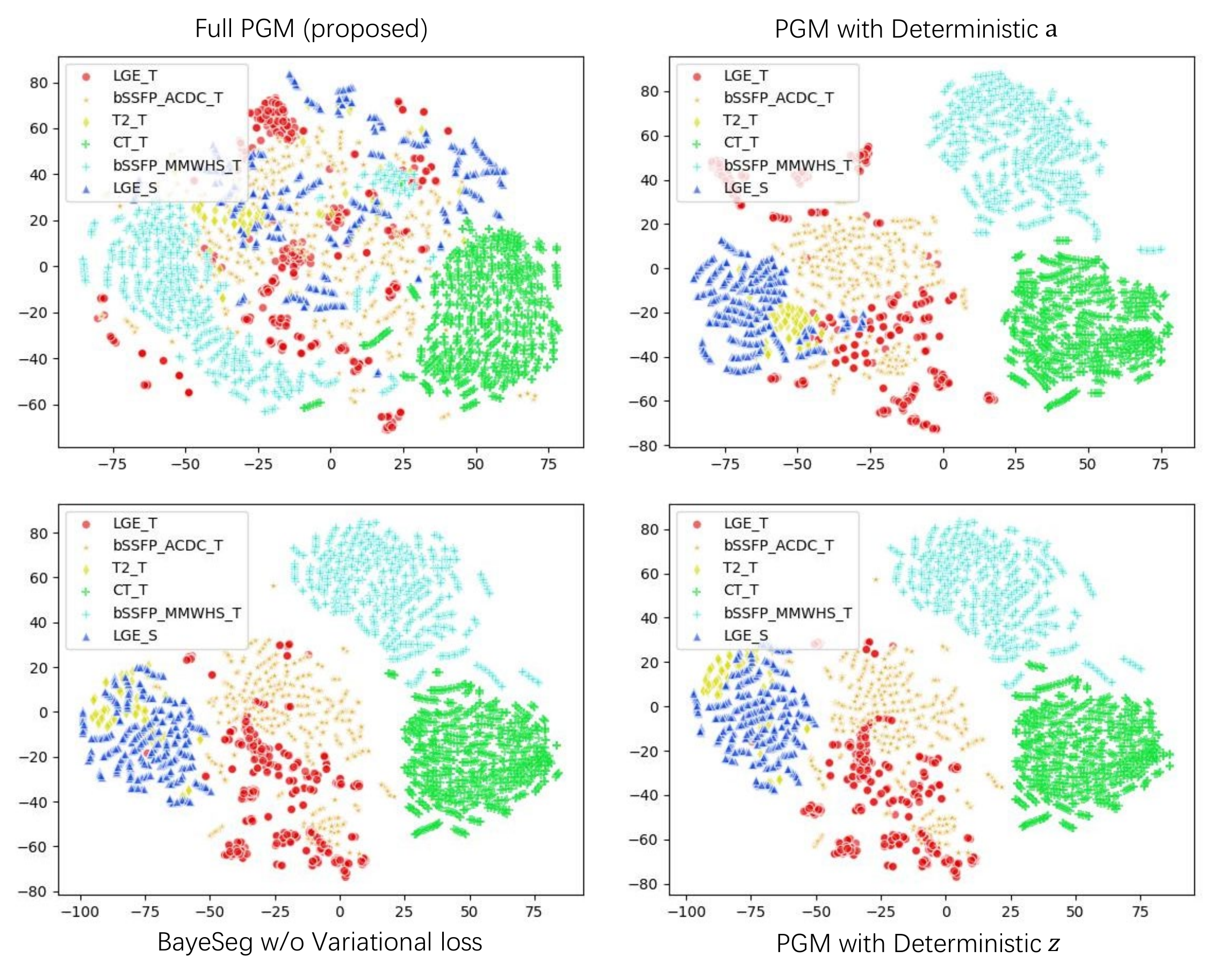}\\[-3ex]
    \caption{t-SNE visualizations of the shape $\bm x$ under different PGM configurations. Here, suffixes S denotes the source domain LGE of MSCMR, while T denotes the target domains.}
    \label{fig:tsne}
\end{figure}

To further verify the benefits of joint modeling, we visualized the shapes extracted by BayeSeg under different PGM settings through t-SNE. As Fig. \ref{fig:tsne} shows, BayeSeg that jointly models the image and label statistics succeeded in centralizing the distributions of $\bm x$ for all target domains. Moreover, the distribution of $\bm x$ extracted on the source domain has greater coverage, which would help BayeSeg generalize better compared to other two incomplete PGM configurations.

\begin{table*}[!t]
\caption{Quantitative average Dice score of different PGM configurations. Here, ``Network'' denotes the stochastic mapping of the corresponding CNN. And ``Loss'' denotes the variational loss that related to the corresponding variable.}\label{tab5:config}
\centering
 \resizebox{1\linewidth}{!}{
\begin{tabular}{|c|cc|cc|c|ccccc|}
\hline
 \multirow{2}{*}{Model}&\multicolumn{2}{c|}{Stochastic $\bm a$}& \multicolumn{2}{c|}{Stochastic $\bm z$} &LGE of MSCMR & T2 of MSCMR  & LGE of EMIDEC & bSSFP of ACDC & bSSFP of MMWHS & CT of MMWHS  \\
 \cline{2-5}
 &Network&Loss&Network&Loss&(Source) & \multicolumn{5}{c|}{(Target)}\\
 \hline
 \#1(proposed)&Y&Y&Y&Y&\textbf{80.9$\pm$6.8}
 &\textbf{63.1$\pm$17.6} 
 &\textbf{67.0$\pm$16.4}
 &\textbf{74.1$\pm$12.0}
 &\textbf{73.5$\pm$14.1}
 &79.4$\pm$10.2
\\
  \hline
 \#2&N&Y&Y&Y   
  &80.1$\pm$8.1
 &12.2$\pm$9.9
 &57.8$\pm$23.0
 &72.8$\pm$14.6
 &67.5$\pm$17.9
 &62.5$\pm$21.5
\\ 
 \#3&Y&N&Y&Y  
  &71.6$\pm$9.6
 &57.2$\pm$14.9
 &46.2$\pm$16.7
 &58.5$\pm$13.7
 &45.1$\pm$14.0
 &45.1$\pm$15.5
\\ 
  \#4&N&N&Y&Y    
  &79.8$\pm$8.3
 &25.2$\pm$16.6
 &60.9$\pm$22.0
 &73.8$\pm$13.7
 &67.3$\pm$18.4
 &73.3$\pm$17.3
\\ 
  \hline
 \#5&Y&Y&N&Y   
 &77.6$\pm$9.6
 &42.4$\pm$18.2
 &64.4$\pm$17.9
 &71.9$\pm$13.3
 &67.5$\pm$17.9
 &62.5$\pm$21.5
\\ 
 \#6&Y&Y&Y&N 
 &79.6$\pm$7.7
 &49.5$\pm$20.4
 &65.4$\pm$17.4
 &73.6$\pm$12.5
 &71.6$\pm$14.4
 &79.1$\pm$10.9
\\ 
\#7&Y&Y&N&N  
 &79.3$\pm$8.4
 &46.6$\pm$23.2
 &64.8$\pm$19.9
 &72.6$\pm$13.1
 &72.8$\pm$14.7
 &\textbf{80.3$\pm$10.0}
\\ 
\hline
\end{tabular}}
\end{table*}

\section{Discussion}
Although the extracted shape can greatly improve model generalizability, it is not completely domain-invariant. The upper left t-SNE in Fig. \ref{fig:tsne} shows that the shape shift between the source domain LGE and the target domain T2 is the smallest one, so the proposed method should achieve the best generalization performance on T2 sequence if the shape is completely domain-invariant. However, the quantitative results in Table \ref{tab5:config} shows that our method delivers the worst generalization ability on the T2 sequence. To explain this phenomenon, we need to rethink the shape shift between medical images. For medical image segmentation, the shape shift comes from two aspects, i.e., anatomy-independent shift and anatomy-relevant shift. The t-SNE is determined by both of them, but the segmentation results are only related to the latter. Since only the orientations between source LGE and target T2 in MSCMR were aligned, the anatomy-independent shift of T2 would be greatly eliminated, making the distribution of T2 the closest to that of LGE in the t-SNE. However, T2 visually has the largest anatomy-relevant shift, making the proposed method present the lowest generalization performance on T2. Therefore, extracting domain-invariant, particularly only anatomy-relevant features is still an open challenge. 
%BayeSeg目前的局限：没有引入因果关系？

Causal deep learning is promising for domain generalization in medical image segmentation. Recently, a few causality-based methods were developed to find an ideal representation that mainly causes the prediction (i.e, structure shape) to reduce the effects of other factors. Commonly, causal relationships are modeled by structural causal models, where the interested factor is fixed. \cite{rw/feature-disentanglement-causality/2021} introduced the latent causal invariant model to explicitly distinguish the causal factors and other factors, achieving remarkable performance on classification. For medical image segmentation, \cite{rw/feature-disentanglement-causality/2022} simulated different possible MRI imaging processes to disturb the image appearance, pursuing the domain-invariance of causal factors. Therefore, imagining unseen scenarios by considering causality in medical imaging is greatly helpful to domain generalization. Such a problem is known as counterfactual inference in statistics, and becomes popular in machine learning recently \citep{CausalML/2022}. In our future work, we will further develop a causal Bayesian framework by rethinking the causality in medical image segmentation.  

\section{Conclusion}
In this work, we have proposed a Bayesian segmentation framework (BayeSeg) through the joint modeling of image and label statistics to promote the interpretability and generalization capability for medical image segmentation.
Specifically, we decompose images into appearance and shape, where hierarchical Bayesian priors are assigned, forcing appearance and shape to model the domain-specific appearance and the domain-stable shape information respectively. After that, the segmentation predictions are generated from shape by modeling of label statistics. 
Additionally, we have presented a variational method of inferring posteriors implemented by neural networks.
Quantitative and qualitative experimental results on prostate segmentation and cardiac segmentation tasks have shown the effectiveness of our proposed method. 
Moreover, we have further interpreted overall posteriors extracted from BayeSeg, and validated the benefits of the explicit prior modeling and the joint modeling of image and label statistics on promoting generalization capability through ablation studies.

\section*{Acknowledgments}
The authors would like to thank Xinzhe Luo and Boming Wang for useful comments and proofread of the manuscript.

%%Harvard
\bibliographystyle{model2-names.bst}\biboptions{authoryear}
\bibliography{refs}

\end{document}